\documentclass[preprint,12pt,authoryear]{elsarticle}

\usepackage{amssymb}
\usepackage{amsmath}

\usepackage{caption}
\usepackage{subcaption}
\usepackage{multirow}
\usepackage{url}
\usepackage{booktabs}
\usepackage[percent]{overpic}
\usepackage{xcolor}
\usepackage{hyperref}

\hypersetup{
    colorlinks=true,
    linkcolor=blue,
    filecolor=magenta,      
    urlcolor=cyan
}

\journal{arXiv}

\begin{document}

\newcommand{\blue}[1]{\textcolor{blue}{#1}}
\newcommand{\black}[1]{\textcolor{black}{#1}}

\begin{frontmatter}



\title{A Step towards Automated and Generalizable Tactile Map Generation using Generative Adversarial Networks} 


\author{David G Hobson}
\ead{davidhobson@cmail.carleton.ca}
\author{Majid Komeili\corref{cor1}} 
\ead{majid.komeili@carleton.ca}

\cortext[cor1]{Corresponding author}

\affiliation{organization={School of Computer Science, Carleton University},
            addressline={1125 Colonel By Dr}, 
            city={Ottawa},
            postcode={K1S 5B6}, 
            state={Ontario},
            country={Canada}}

\begin{abstract}
Blindness and visual impairments affect many people worldwide. For help with navigation, people with visual impairments often rely on tactile maps that utilize raised surfaces and edges to convey information through touch. Although these maps are helpful, they are often not widely available and current tools to automate their production have similar limitations including only working at certain scales, for particular world regions, or adhering to specific tactile map standards. To address these shortcomings, we train a proof-of-concept model as a first step towards applying computer vision techniques to help automate the generation of tactile maps. We create a first-of-its-kind tactile maps dataset of street-views from Google Maps spanning 6500 locations and including different tactile line- and area-like features.\footnote{Dataset available at \url{https://doi.org/10.20383/103.0797}} Generative adversarial network (GAN) models trained on a single zoom successfully identify key map elements, remove extraneous ones, and perform inpainting with median F1 and intersection-over-union (IoU) scores of better than 0.97 across all features. Models trained on two zooms experience only minor drops in performance, and generalize well both to unseen map scales and world regions. Finally, we discuss future directions towards a full implementation of a tactile map solution that builds on our results.
\end{abstract}


\end{frontmatter}

\section{Introduction} \label{sec:intro}
An estimated 253 million people worldwide are visually impaired according to the World Health Organization \citep{who_paper}. Among people with visual impairments (PVI), navigation is a particularly challenging area in their lives, but many find tactile maps, which utilize raised surfaces and edges to convey information through touch, to be very helpful \citep{accessibility, shabiralyani}. Tactile maps improve autonomy and quality of life for PVI \citep{espinosa, jacobson}, while also helping them be more comfortable and confident in new environments \citep{rowell_feeling_your_way, rowell}. Despite their usefulness and the positive attitudes of users towards them, their adoption is not widespread \citep{user_study}. 

A major factor contributing to this limited availability is the fact that designing tactile maps can be a complicated and multifaceted process \citep{lobben}. Visual maps can contain a lot of information in relatively small areas, but since tactile maps need to be simple enough to remain legible by touch, a trade-off is necessary to reduce map complexity while still maintaining as many important elements as possible \citep{way}. As such, designing tactile maps often requires a tactile designer, which can make them expensive or subject to long processing times \citep{lobben}. While modern technologies like 3D printers enable tactile maps to be printed from home, doing so often involves specialized software or technical expertise \citep{tmc, miele, stampach}. 

To address these issues, several works develop software or other system solutions to automate the creation of tactile maps. The TMAP service, for example, allows tactile maps to be generated from street-views on OpenStreetMaps which can then be printed using a 3D printer \citep{tmap}. Other solutions leverage GIS databases \citep{mapy, tmacs}, or national mapping data \citep{stampach}, while other works have created tactile mapping tools for building interiors \citep{tmc}.

\begin{figure}[t]
    \centering
    \includegraphics[width=\textwidth]{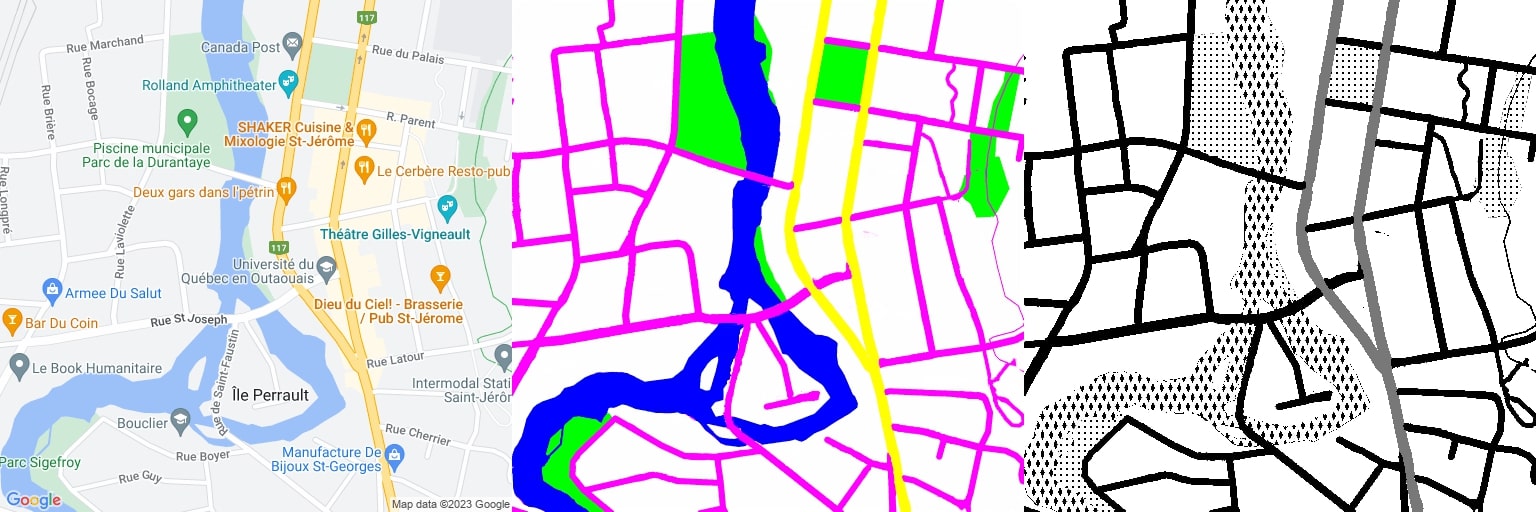}
    \caption{Source (left), tactile (middle), and a possible textured tactile map (right). These correspond to the model input, model output, and a post-processed model output.}
    \label{fig:dataset_ex}
\end{figure}

Although these efforts all greatly contribute to improving the accessibility of real-world tactile maps, they all have similar shortcomings. While many of the tools work for different scales, they do not generalize to scales outside of normal street-viewing, such as focusing on particular landmarks, views of cities, or states/provinces. 
The tools are also often optimized to specific world regions, and adhere to different tactile standards and guidelines \citep{stampach, wabinski}. Finally, these solutions are all specific to tactile mapping and do not generalize to creating other kinds of tactile graphics or images. 

A new approach is the use of artificial intelligence (AI) and computer vision (CV) to automate the generation of tactile graphics. These methods offer considerable potential for improving the generalizability, speed, and flexibility of current solutions, while offering the possibility of creating general-purpose image-to-tactile models capable of generating tactile graphics across several domains. 

At this point, however, very little research explores this area. \citet{heydari} employ such methods for the generation of tactile plots and graphs, being the only major effort to date. In general, a lack of large-scale training datasets is a major obstacle for research and applications in this field. 
More broadly, complete CV-based solutions for generating tactile maps involve multiple components. Models need to perform image processing to identify relevant features, text processing to identify relevant text, text conversion Braille, among other tasks. Even constrained to the image processing step, there is a need to identify features of relevance to PVI, remove other extraneous features, and fill in the gaps of those removed features. Moreover, models need the ability to generalize to new settings, including different map scales and world regions with different languages and map standards. 

In light of these questions, we provide a first step towards addressing these problems by applying CV-based approaches to tactile map generation for the first time. Specifically, we limit ourselves to the image processing step. We contribute a new first-of-its-kind dataset of visual-tactile map pairs of street-views from Google Maps covering a total of six line- and area-like features. We then use our dataset to train a proof-of-concept image-to-image generation model to convert visual maps to their tactile map equivalents. We show that Pix2Pix-based generative adversarial network (GANs) models are capable of performing the tasks most necessary for tactile map generation, and demonstrate that these models can interpolate and extrapolate effectively to different scales, as well as handle text from unseen languages.

The following section provides a brief overview of AI-related applications to improving accessibility for PVI, and briefly summarizes the work by \citet{heydari} on tactile plot generation. Subsequent sections describe the methodology, including the creation of the dataset and the experimental details, and finally the results of the trained models are presented.

\section{Related Work} \label{sec:related_work}
Overall, the application of AI techniques towards improving accessibility for PVI is very limited. Research includes the development of touchscreen systems with environmental perception \citep{yimin}, systems to automate video descriptions \citep{yuksel1, yuksel2}, and the creation of assistive tools for understanding images in text \citep{ganesan}. More closely related to our work, \citet{bose} uses CV techniques are classify tactile images, and both \citet{gonzalez} and \citet{martinez} develop classifiers for identifying the suitability of an image for conversion into a tactile graphic. As noted previously, only \citet{heydari} applies CV techniques directly to the generation of tactile graphics.

Specifically, \citet{heydari} treats the generation problem as an image-to-image translation task and employs the Pix2Pix-GAN architecture \citep{pix2pix} to create tactile versions of 2D plots, including bezier curves, polygons, scatter plots, and bar charts. They create a synthetic dataset of randomly generated plots where each input (the source plot) is an RGB image of the visual plot, and the output is an image representation of the corresponding tactile plot. These pairs of visual and tactile plots are referred to as source-tactile pairs which is the terminology we adopt here. For the format of the output tactile plots, one of the representations includes an RGB image where different tactile components were represented by unique colors. Figure \ref{fig:dataset_ex} shows an example when applied to our map use-case with the leftmost image representing the input/source map, and the middle image being the simplified tactile map in its RGB representation. Although the outputs do not visually appear like a physical tactile map, these tactile representations are useful because they are agnostic to any specific tactile pattern scheme. This offers the advantage that users can choose any desired tactile pattern scheme in later downstream processing, while eliminating the need to retrain different models for different tactile patterns. The rightmost image of Figure \ref{fig:dataset_ex} illustrates this, where each of the tactile features is assigned an arbitrary texture.

With this setup, the model key tasks are to identify and color the key plot components, simplify the plot by removing extraneous elements, and inpaint across to produce the final tactile output. With these being variants of semantic segmentation, sketch generation, and inpainting tasks, \citet{heydari} uses the Pix2Pix architecture for general purpose image-to-image translation. The models achieve good results achieving 0.0037 and 0.04 foreground and background mean squared error, respectively.  

Motivated by these results, we apply the same model architecture to the more challenging task of tactile map generation. Like in \citet{heydari}, we train models to identify and color features from a source map, however now from images featuring a larger number of classes with much greater visual complexity. Since map features such as icons, symbols, and text are extraneous for PVI, we train the models to remove these elements from the tactile outputs. Although the preservation of certain icons, and the identification and conversion of map text to Braille are important considerations, we leave these tasks as future work. 

\section{Methodology} \label{sec:methodology}
The following subsections provide details on the task definition, the dataset and its creation, the model architecture, evaluation criteria, and experiments performed.

\subsection{Task Definition}

Following \citet{heydari}, we treat the task of tactile map generation as an image-to-image translation task. The input to the model is an RGB image of a visual map and the output is an RGB image representation of the corresponding tactile maps with each tactile element represented by a unique color. We refer to this pairing as a source-tactile map pair, with an example shown in Figure \ref{fig:dataset_ex}. In this paper, we focus on maps of street-views. 

The model's tasks are to identify and simplify the features most relevant to PVI, remove extraneous elements like icons and street names, and inpaint the resulting discontinuities. Models should also be able to generalize to different map scales and to different world regions.

Importantly, the goal in generating RGB representations of the tactile maps is to ensure that the outputs are agnostic to any particular tactile mapping standards, as this allows the model to be applicable regardless of the specific regional tactile guideline being followed during printing.  
Using our representations, however, the choice of physical textures, line widths, and other considerations can be fulfilled during post-processing according to the desired guideline. For a more detailed discussion on tactile guidelines, including map legibility, see \ref{sec:appendix_further_discussion} and \ref{sec:appendix_map_legibility}.

\subsection{Dataset}

Since no large-scale dataset of tactile street maps are available, we create a custom, first-of-its-kind dataset using the Google Maps Static API (GMS API). We chose Google Maps since it is familiar to the general public and possesses broad coverage of many world regions.

In total, we collect source-tactile map pairs from 6500 locations at four different scales: 5500 are from four English speaking countries (namely Australia, Canada, the United Kingdom, and the United States), and consist of cities, landmarks, hospitals, and universities and colleges. We split these locations into training and test sets of 5000 and 500 image pairs, respectively. The additional 1000 map pairs are for world cities outside of the aforementioned countries, and are used to test the models' abilities to generalize to unseen regions and languages. We refer to these two test sets as the \textit{English} and \textit{World} test sets accordingly. For the full details on the location selection and breakdown, see \ref{sec:appendix_loc_details}.

All data from this project is made publicly available through the Federated Research Data Repository \citep{dataset}.

\subsubsection{Map Features}
In choosing the map features to include in the tactile maps, we select a total of six line- and area-like features as recommended by \citet{regis}. We chose the features to maximize usefulness for PVI while ensuring the features are still visually distinguishable on the source maps. They include streets (colored in magenta on the tactile maps), buildings (colored in cyan), parks and green spaces (green), and bodies of water (blue); all of which have been identified as important in user studies \citep{rowell_feeling_your_way, stampach}. The ground truth column of Figure \ref{fig:good_16_18} shows examples. Highways (yellow) and medical facilities (grey) are also chosen to probe the models' abilities to identify features with similar shapes (to roads and buildings, respectively), but different colors (Google uses pink to signify medical areas but usually grey shades to show normal buildings). 

While point features, including bus stops and traffic lights, are also relevant to PVI \citep{papadopoulos}, the identification of these features is left as future work. \ref{sec:appendix_further_discussion} offers an overview of research studying map features relevant to PVI. The breakdown of the classes in the dataset can be found in \ref{sec:appendix_dataset_breakdown}.

\subsubsection{Data Collection and Map Scales}

We use the GMS API to collect all maps, both source and tactile. The source maps consist of the original, unaltered maps from Google. For the tactile maps, we apply an API style to color the features of interest and remove all labels, icons, text, and other symbols. While some of icons and labels may be important for PVI, in general, they interfere with the legibility of the tactile map and therefore are removed. The source map dimensions are 512x512 and the tactile maps at 572x572. The latter is chosen so that the Google logo and copyright signatures can be removed in the center-cropped 512x512 images, which is done prior to training.

For map scale, Google Maps uses a custom zoom-level score ranging from 1 to 19, with 1 being the most zoomed out and 19 being the most zoomed in. Zoom levels are only integer-valued and buildings are only visible from zoom 17 and above. For proper street viewing, zooms between 15 and 19 are most suitable. For this research, we chose to collect maps at zoom levels of 15, 16, 17, and 18 to include maps both with and without buildings. 

All calls to the GMS API specify the location name as the map center center and used a zoom between 15 and 18. We collect a complete dataset (6500 source-tactile pairs) for each zoom from 15 to 18 (inclusive) for a total of four datasets. For the complete details on the API style specification, see \ref{sec:appendix_api}.

\subsection{Model} \label{sec:model}

We employ the same model architecture as that from \citet{heydari}, namely the Pix2Pix-GAN architecture \citep{pix2pix} using a UNet++ generator \citep{unet++} and a PatchGAN \citep{patchgan} discriminator.  While other architectures can be used, the success of the Pix2Pix architecture, as well as UNet++, to a broad range of image-to-image translation tasks motivates our use of them here \citep{unet_success, bollepalli_pix2pix, kim_pix2pix, segmentation_cell, wijanto_pix2pix}.

The loss function consists of a distance-based L$_{1}$-loss and a GAN loss as in the original Pix2Pix model. We use the same default hyperparameters as \citet{heydari}.

\subsection{Experiments} \label{sec:exps}

Three models are trained and evaluated. We train two models on a single zoom at each of levels 16 and 18, which serve as the performance baselines. The other model is trained on both the zoom 16 and 18 training sets together. We refer to these models as the Zoom-16, Zoom-18, and Zoom-16/18 models, respectively. The Zoom-16/18 model is also called the double zoom model. These zooms are chosen since they contain maps both with and without buildings (zoom 18 and zoom 16, respectively). This allows for the Zoom-16/18 model to be evaluated on both these scenarios, and see how well it can interpolate to zoom 17 and extrapolate to zoom 15. We leave extrapolation to zoom 19 as future work since maps can contain buildings with many 3D affects creating a more advanced task. 

Models are evaluated on the aggregated semantic segmentation, sketch generation, and inpainting tasks, and on both the \textit{English} and \textit{World} test sets to indicate performance at different scales, and unseen countries and regions. Models are only tested on their trained zooms levels and on compatible zoom levels. This includes zoom 15 for the Zoom-16 model (both do not contain buildings), and zoom 17 for the Zoom-18 model (both contain buildings). The Zoom-16/18 model is evaluated on the test sets at all zooms.

We evaluate using intersection-over-union (IoU), F1, precision, and recall over all six tactile classes. To apply these metrics, we first segment pixels in the ground truth and predicted RGB images into classes by assigning them to their closest color based on the L$_{1}$ distance. To reduce misclassifications into the grey hospital class, however, if the distance from a pixel to its closest color is greater than a threshold of 230, it is classified as background.

\ref{sec:appendix_training_details} provides additional details on the model training and data augmentation.

\section{Results} \label{sec:results}

\begin{table}[t!]
\centering
\resizebox{\textwidth}{!}{
    \begin{tabular}{ll||ccc|ccc|ccc|ccc}
    \toprule
    & \textbf{Metric} & \multicolumn{3}{c|}{IoU} & \multicolumn{3}{c|}{F1} & \multicolumn{3}{c|}{Precision} & \multicolumn{3}{c}{Recall} \\
    & \textbf{Model} & Double & Single & \emph{Diff} & Double & Single & \emph{Diff} & Double & Single & \emph{Diff} & Double & Single & \emph{Diff} \\
    \hline
    \textbf{Class} & \textbf{Statistic} &  &  &  &  &  &  &  &  &  &  &  &  \\
    \midrule
    \multirow[c]{2}{*}{Streets} & median & 94.8 & 97.1 & -2.3 & 97.3 & 98.5 & -1.2 & 96.3 & 98.8 & -2.5 & 98.5 & 98.3 & 0.2 \\
     & mean & 94.2 & 96.4 & -2.2 & 97.0 & 98.1 & -1.1 & 96.1 & 98.4 & -2.3 & 97.9 & 97.9 & 0.0 \\
    \multirow[c]{2}{*}{Highways} & median & 96.7 & 97.2 & -0.5 & 98.3 & 98.6 & -0.3 & 97.7 & 98.6 & -0.9 & 99.1 & 98.7 & 0.4 \\
     & mean & 94.3 & 95.2 & -0.9 & 96.7 & 97.3 & -0.6 & 97.0 & 97.8 & -0.8 & 96.8 & 97.0 & -0.2 \\
    \multirow[c]{2}{*}{Parks} & median & 97.9 & 98.3 & -0.4 & 98.9 & 99.1 & -0.2 & 98.9 & 99.2 & -0.3 & 98.9 & 99.2 & -0.3 \\
     & mean & 95.8 & 96.3 & -0.5 & 97.3 & 97.6 & -0.3 & 98.0 & 98.2 & -0.2 & 97.4 & 97.7 & -0.3 \\
    \multirow[c]{2}{*}{Water} & median & 98.3 & 98.5 & -0.2 & 99.1 & 99.2 & -0.1 & 99.1 & 99.2 & -0.1 & 99.4 & 99.5 & -0.1 \\
     & mean & 97.2 & 97.4 & -0.2 & 98.5 & 98.7 & -0.2 & 98.4 & 98.5 & -0.1 & 98.7 & 98.9 & -0.2 \\
    \multirow[c]{2}{*}{Buildings} & median & N/A & N/A & N/A & N/A & N/A & N/A & N/A & N/A & N/A & N/A & N/A & N/A \\  
     & mean & N/A & N/A & N/A & N/A & N/A & N/A & N/A & N/A & N/A & N/A & N/A & N/A \\
    \multirow[c]{2}{*}{Hospitals} & median & 98.7 & 98.9 & -0.2 & 99.3 & 99.4 & -0.1 & 99.2 & 99.5 & -0.3 & 99.4 & 99.4 & 0.0 \\
     & mean & 95.2 & 95.6 & -0.4 & 97.0 & 97.3 & -0.3 & 97.5 & 97.4 & 0.1 & 96.8 & 97.6 & -0.8 \\
     \bottomrule
    \end{tabular}
}
\caption{Class performances on the \textit{English} zoom-16 test set.}
\label{tab:zoom_16_results}
\end{table}

\begin{table}
\centering
\resizebox{\textwidth}{!}{
    \begin{tabular}{cc||ccc|ccc|ccc|ccc}
    \toprule
    & \textbf{Metric} & \multicolumn{3}{c|}{IoU} & \multicolumn{3}{c|}{F1} & \multicolumn{3}{c|}{Precision} & \multicolumn{3}{c}{Recall} \\
    & \textbf{Model} & Double & Single & \emph{Diff} & Double & Single & \emph{Diff} & Double & Single & \emph{Diff} & Double & Single & \emph{Diff} \\
    \hline
    \textbf{Class} & \textbf{Statistic} &  &  &  &  &  &  &  &  &  &  &  &  \\
    \midrule
    \multirow[c]{2}{*}{Streets} & median & 97.4 & 97.1 & 0.3 & 98.7 & 98.5 & 0.2 & 98.2 & 98.4 & -0.2 & 99.3 & 98.9 & 0.4 \\ 
     & mean & 96.4 & 95.4 & 1.0 & 98.1 & 97.5 & 0.6 & 97.7 & 97.8 & -0.1 & 98.6 & 97.5 & 1.1 \\
    \multirow[c]{2}{*}{Highways} & median & 98.4 & 98.0 & 0.4 & 99.2 & 99.0 & 0.2 & 98.8 & 98.6 & 0.2 & 99.7 & 99.6 & 0.1 \\   
     & mean & 97.3 & 96.6 & 0.7 & 98.2 & 97.8 & 0.4 & 97.8 & 97.5 & 0.3 & 98.6 & 98.1 & 0.5 \\
    \multirow[c]{2}{*}{Parks} & median & 98.8 & 98.3 & 0.5 & 99.4 & 99.1 & 0.3 & 99.4 & 99.3 & 0.1 & 99.4 & 99.3 & 0.1 \\   
     & mean & 95.8 & 95.6 & 0.2 & 97.0 & 96.9 & 0.1 & 97.1 & 96.9 & 0.2 & 97.0 & 97.0 & 0.0 \\
    \multirow[c]{2}{*}{Water} & median & 99.3 & 99.1 & 0.2 & 99.6 & 99.5 & 0.1 & 99.7 & 99.7 & 0.0 & 99.8 & 99.6 & 0.2 \\    
     & mean & 98.7 & 97.0 & 1.7 & 99.3 & 98.4 & 0.9 & 99.3 & 99.3 & 0.0 & 99.4 & 97.6 & 1.8 \\
    \multirow[c]{2}{*}{Buildings} & median & 97.6 & 97.3 & 0.3 & 98.8 & 98.6 & 0.2 & 98.5 & 98.1 & 0.4 & 99.2 & 99.4 & -0.2 \\ 
     & mean & 96.3 & 95.5 & 0.8 & 97.9 & 97.5 & 0.4 & 97.8 & 96.6 & 1.2 & 98.2 & 98.7 & -0.5 \\
    \multirow[c]{2}{*}{Hospitals} & median & 98.8 & 98.8 & 0.0 & 99.4 & 99.4 & 0.0 & 99.5 & 99.5 & 0.0 & 99.4 & 99.3 & 0.1 \\   
     & mean & 98.4 & 98.3 & 0.1 & 99.2 & 99.1 & 0.1 & 99.4 & 99.3 & 0.1 & 99.0 & 99.0 & 0.0 \\
     \bottomrule
    \end{tabular}
}
\caption{Class performances on the \textit{English} zoom-18 test set.}
\label{tab:zoom_18_results}
\end{table}

\begin{figure}[ht!]
    \centering 

    \begin{subfigure}[b]{0.23\textwidth}
        \centering
        \includegraphics[width=\textwidth]{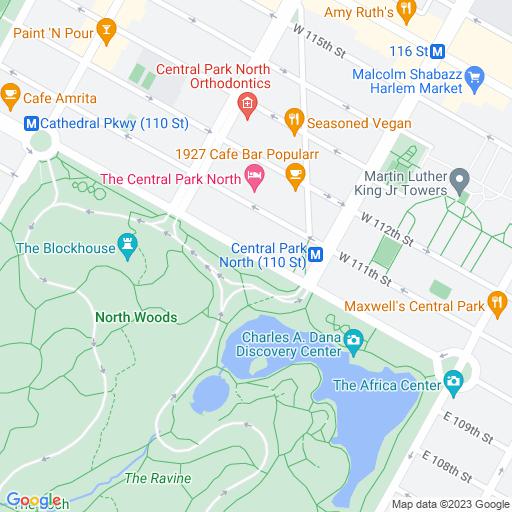}
    \end{subfigure}
    \begin{subfigure}[b]{0.23\textwidth}
        \centering
        \includegraphics[width=\textwidth]{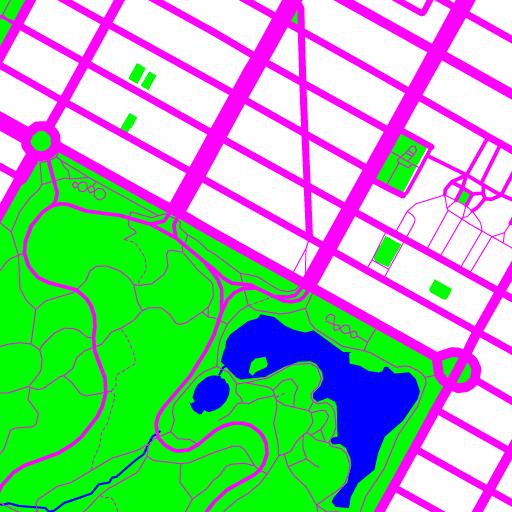}
    \end{subfigure}
    \begin{subfigure}[b]{0.23\textwidth}
        \centering
        \includegraphics[width=\textwidth]{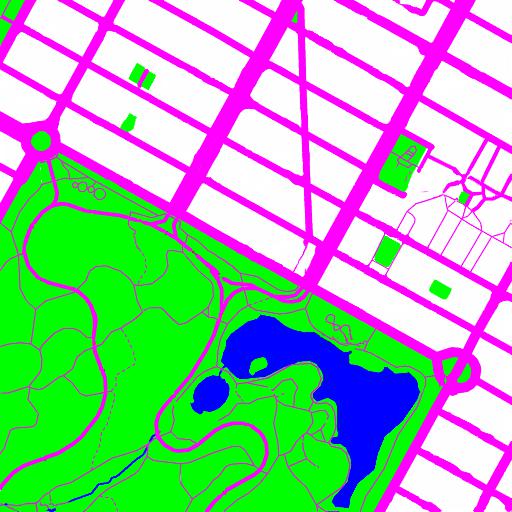}
    \end{subfigure}
    \begin{subfigure}[b]{0.23\textwidth}
        \centering
        \includegraphics[width=\textwidth]{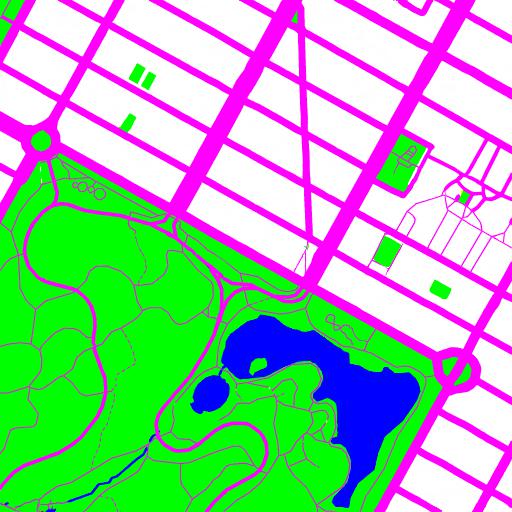}
    \end{subfigure}

    \vspace{4pt}

    \begin{subfigure}[b]{0.23\textwidth}
        \centering
        \includegraphics[width=\textwidth]{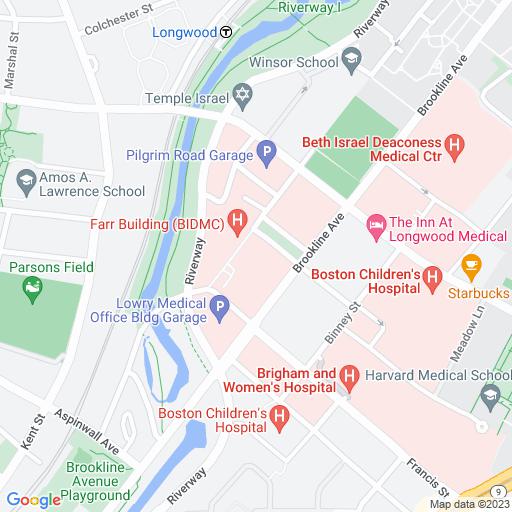}
    \end{subfigure}
    \begin{subfigure}[b]{0.23\textwidth}
        \centering
        \includegraphics[width=\textwidth]{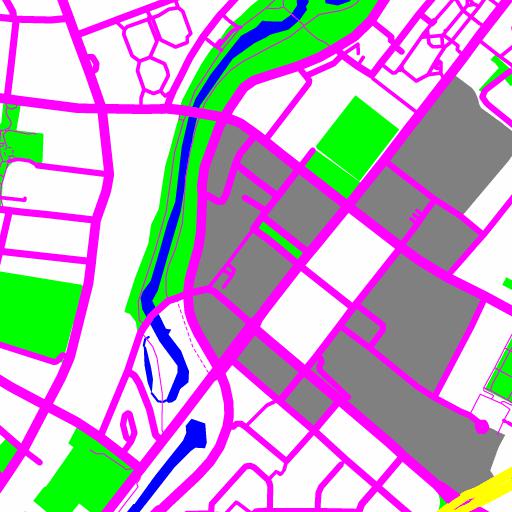}
    \end{subfigure}
    \begin{subfigure}[b]{0.23\textwidth}
        \centering
        \includegraphics[width=\textwidth]{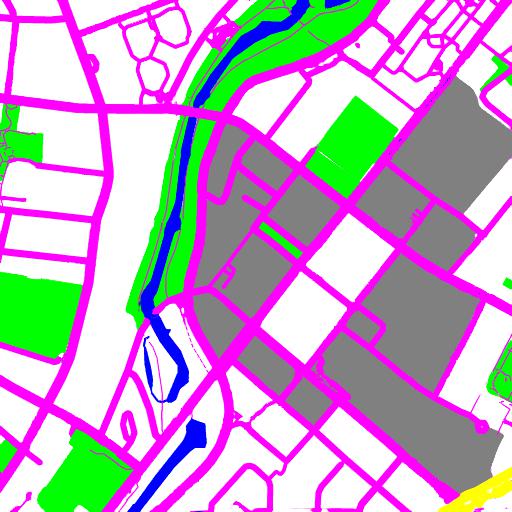}
    \end{subfigure}
    \begin{subfigure}[b]{0.23\textwidth}
        \centering
        \includegraphics[width=\textwidth]{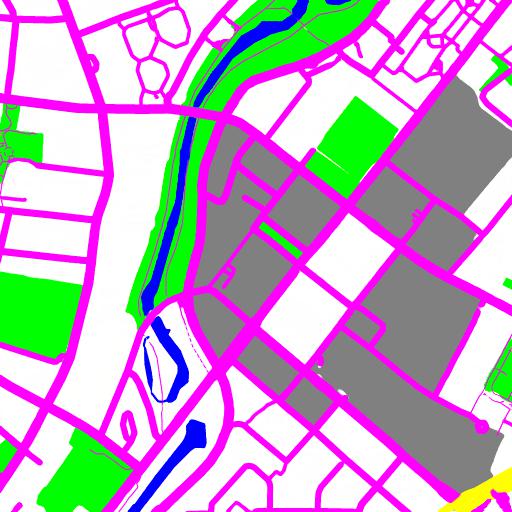}
    \end{subfigure}

    \vspace{4pt}

   \begin{subfigure}[b]{0.23\textwidth}
        \centering
        \includegraphics[width=\textwidth]{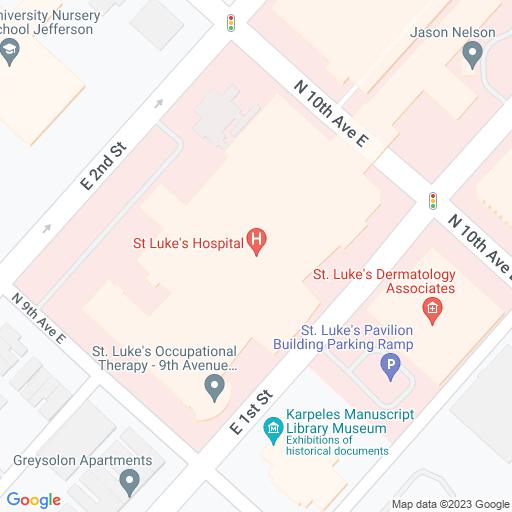}
    \end{subfigure}
    \begin{subfigure}[b]{0.23\textwidth}
        \centering
        \includegraphics[width=\textwidth]{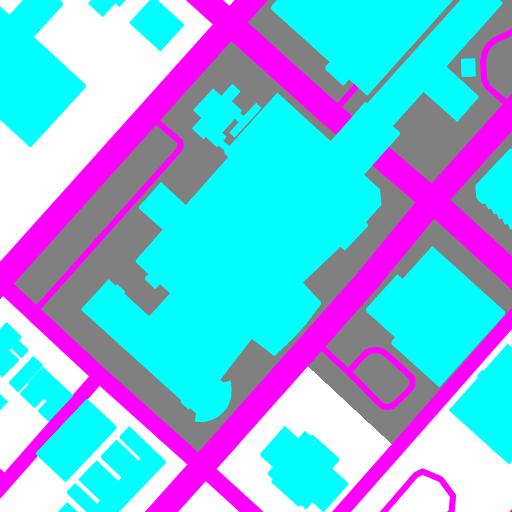}
    \end{subfigure}
    \begin{subfigure}[b]{0.23\textwidth}
        \centering
        \includegraphics[width=\textwidth]{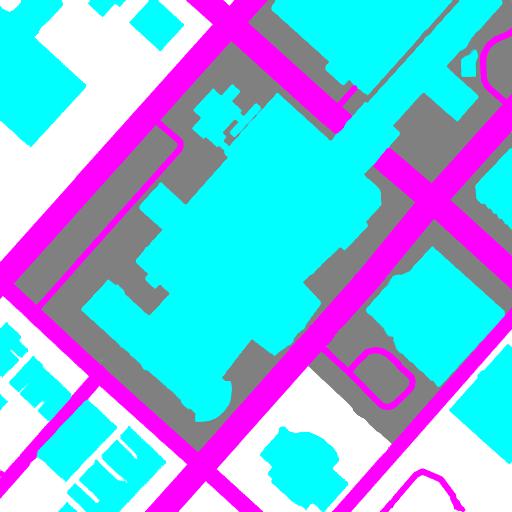}
    \end{subfigure}
    \begin{subfigure}[b]{0.23\textwidth}
        \centering
        \includegraphics[width=\textwidth]{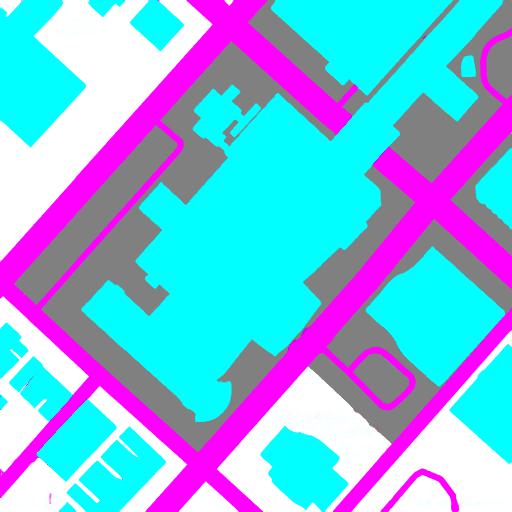}
    \end{subfigure}

    \vspace{4pt}

    \begin{subfigure}[b]{0.23\textwidth}
        \centering
        \includegraphics[width=\textwidth]{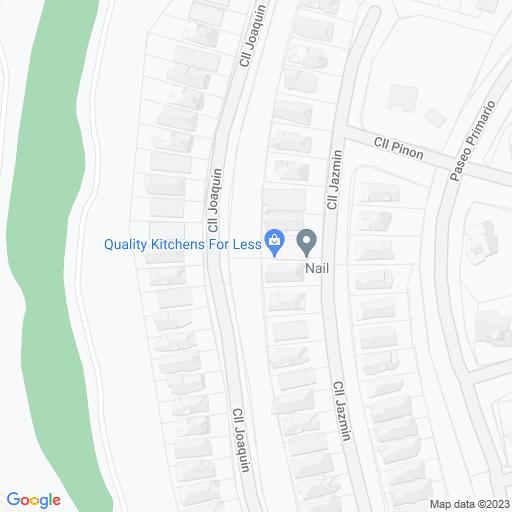}
    \end{subfigure}
    \begin{subfigure}[b]{0.23\textwidth}
        \centering
        \includegraphics[width=\textwidth]{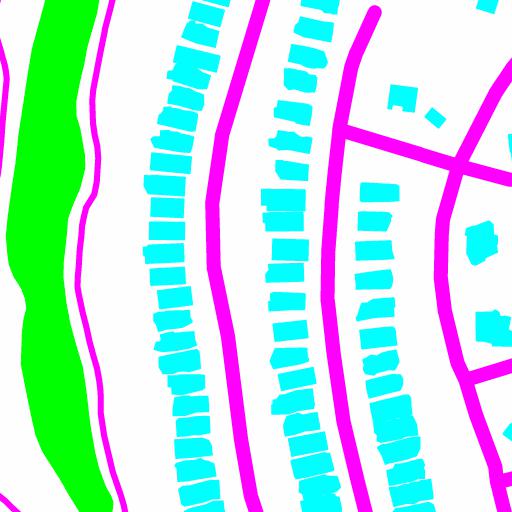}
    \end{subfigure}
    \begin{subfigure}[b]{0.23\textwidth}
        \centering
        \includegraphics[width=\textwidth]{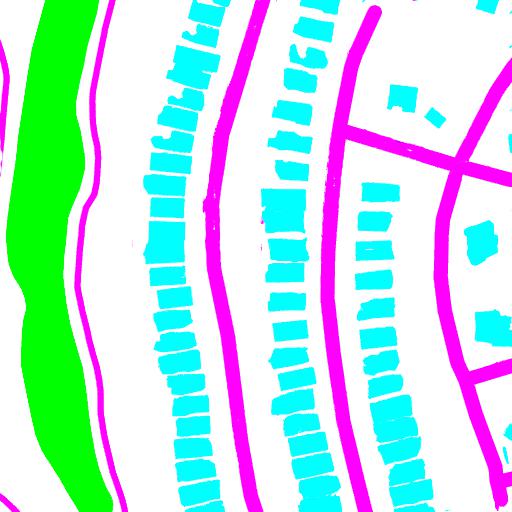}
    \end{subfigure}
    \begin{subfigure}[b]{0.23\textwidth}
        \centering
        \includegraphics[width=\textwidth]{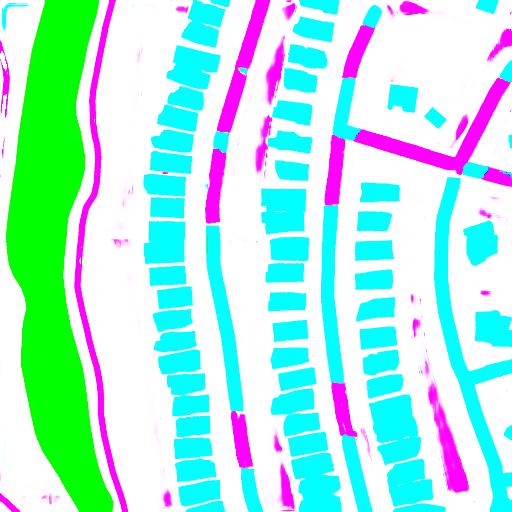}
    \end{subfigure}

    \begin{subfigure}[b]{0.23\textwidth}
        \centering
        \includegraphics[width=\textwidth]{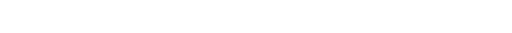}
        \caption{Source Map}
        \label{fig:results_src}
    \end{subfigure}
    \begin{subfigure}[b]{0.23\textwidth}
        \centering
        \includegraphics[width=\textwidth]{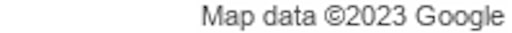}
        \caption{Ground Truth}
        \label{fig:results_gt}
    \end{subfigure}
    \begin{subfigure}[b]{0.23\textwidth}
        \centering
        \includegraphics[width=\textwidth]{Filler_horizontal.jpg}
        \caption{Double Zoom Model}
        \label{fig:results_double}
    \end{subfigure}
    \begin{subfigure}[b]{0.23\textwidth}
        \centering
        \includegraphics[width=\textwidth]{Filler_horizontal.jpg}
        \caption{Single Zoom Model}
        \label{fig:results_single}
    \end{subfigure}
    \caption{Sample model outputs from the zoom 16 and 18 \textit{English} test sets. The copyright underneath the second column applies to all images in that column.}
    \label{fig:good_16_18}
\end{figure}

Tables \ref{tab:zoom_16_results} and \ref{tab:zoom_18_results} compare the results of the double and single zoom models (under the Double and Single column headings, respectively) when applied on the zoom-16 and zoom-18 \textit{English} test sets respectively. We show both median and mean values since the presence of outliers had a tendency to lower the means compared to the medians. Figure \ref{fig:good_16_18} shows some sample outputs with each row representing an example. The source maps are on the left, followed by the ground truth tactile image, the prediction from the Zoom-16/18 model, and the prediction from the single zoom model. The first two rows are from maps at zoom 16 and the last two rows are from zoom 18.

At both zooms, the single and double zoom models achieve very strong performance across all classes with IoU scores exceeding 94\%. As seen in Figure \ref{fig:good_16_18}, the models are successful at removing icons, symbols, and text, including building and street names, and perform inpainting of removed elements effectively. In \black{row 1} of Figure \ref{fig:good_16_18}, the models also exhibit good ability at identifying finer elements like pedestrian paths, with only minor errors associated with inpainting. Row 3 also shows the models had no difficulties with buildings and other features within medical areas.

The \emph{Diff} column shows the difference between the double and single zoom metrics. The Zoom-16/18 model performs well overall, with only slight drops in performance compared to the Zoom-16 model and with slight improvements over the Zoom-18 model. In particular, at zoom 18, the increases in the mean values show improved performance on outliers compared to the Zoom-18 model. For streets and buildings, this is partially due to the data augmentation with the double zoom model, which enables it to better distinguish between the two classes. This is illustrated best in \black{row 4} of Figure \ref{fig:good_16_18}. While the Zoom-16/18 model also improves on the mean metrics over the other classes, this is due to the higher average representation of each class in the maps of the zoom 16 training set compared to the zoom 18 training set.

\subsection{Interpolation and Extrapolation to Different Scales}

\begin{table}[t!]
\centering
\resizebox{\textwidth}{!}{
    \begin{tabular}{cc||ccc|ccc|ccc|ccc}
    \toprule
    & \textbf{Metric} & \multicolumn{3}{c|}{IoU} & \multicolumn{3}{c|}{F1} & \multicolumn{3}{c|}{Precision} & \multicolumn{3}{c}{Recall} \\
    & \textbf{Model} & Double & Single & \emph{Diff} & Double & Single & \emph{Diff} & Double & Single & \emph{Diff} & Double & Single & \emph{Diff} \\
    \hline
    \textbf{Class} & \textbf{Statistic} &  &  &  &  &  &  &  &  &  &  &  &  \\
    \midrule
    \multirow[c]{2}{*}{Streets} & median & 88.7 & 90.3 & -1.6 & 94.0 & 94.9 & -0.9 & 91.5 & 92.5 & -1.0 & 97.0 & 97.6 & -0.6 \\    
     & mean & 87.8 & 89.6 & -1.8 & 93.4 & 94.4 & -1.0 & 91.2 & 92.2 & -1.0 & 95.9 & 96.9 & -1.0 \\
    \multirow[c]{2}{*}{Highways} & median & 91.8 & 91.9 & -0.1 & 95.7 & 95.8 & -0.1 & 94.2 & 93.2 & 1.0 & 98.3 & 99.2 & -0.9 \\
     & mean & 90.2 & 90.6 & -0.4 & 94.7 & 95.0 & -0.3 & 93.2 & 92.4 & 0.8 & 96.6 & 97.9 & -1.3 \\
    \multirow[c]{2}{*}{Parks} & median & 95.1 & 95.8 & -0.7 & 97.5 & 97.8 & -0.3 & 97.5 & 97.7 & -0.2 & 97.6 & 98.0 & -0.4 \\    
     & mean & 92.5 & 93.2 & -0.7 & 95.6 & 96.0 & -0.4 & 96.1 & 96.5 & -0.4 & 96.0 & 96.5 & -0.5 \\
    \multirow[c]{2}{*}{Water} & median & 96.3 & 97.0 & -0.7 & 98.1 & 98.5 & -0.4 & 97.5 & 98.1 & -0.6 & 99.2 & 99.1 & 0.1 \\    
     & mean & 94.3 & 94.9 & -0.6 & 97.0 & 97.3 & -0.3 & 96.0 & 96.6 & -0.6 & 98.1 & 98.1 & 0.0 \\
    \multirow[c]{2}{*}{Buildings} & median & N/A & N/A & N/A & N/A & N/A & N/A & N/A & N/A & N/A & N/A & N/A & N/A \\    
     & mean & N/A & N/A & N/A & N/A & N/A & N/A & N/A & N/A & N/A & N/A & N/A & N/A \\
    \multirow[c]{2}{*}{Hospitals} & median & 91.8 & 92.1 & -0.3 & 95.7 & 95.9 & -0.2 & 95.9 & 96.2 & -0.3 & 95.4 & 95.6 & -0.2 \\    
     & mean & 82.9 & 81.9 & 1.0 & 89.0 & 87.9 & 1.1 & 91.9 & 90.4 & 1.5 & 87.1 & 86.3 & 0.8 \\
     \bottomrule
    \end{tabular}
}
\caption{Class performances on the \textit{English} zoom-15 test set.}
\label{tab:zoom_15_results}
\end{table}

\begin{figure}[t!]
    \centering 
    
   \begin{subfigure}[b]{0.23\textwidth}
        \centering
        \includegraphics[width=\textwidth]{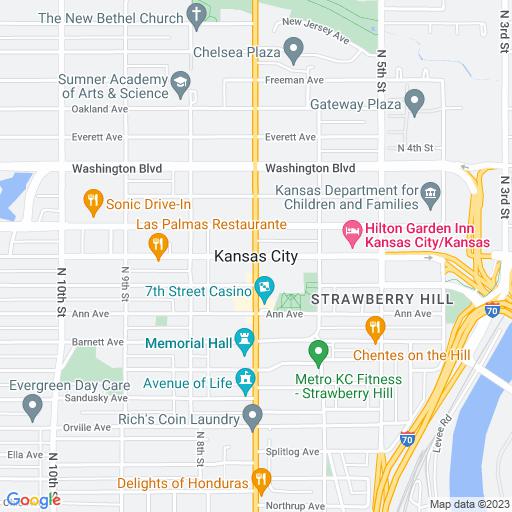}
    \end{subfigure}
    \begin{subfigure}[b]{0.23\textwidth}
        \centering
        \includegraphics[width=\textwidth]{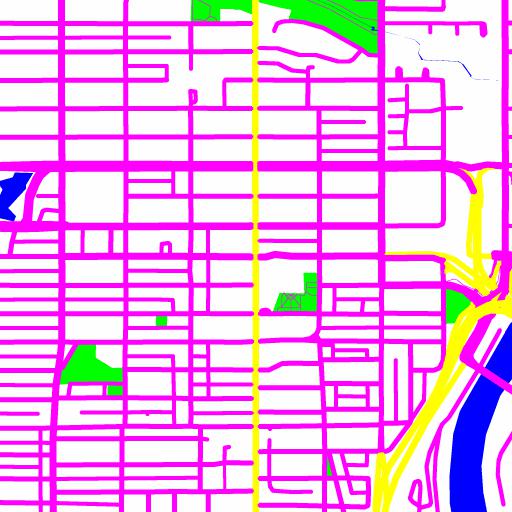}
    \end{subfigure}
    \begin{subfigure}[b]{0.23\textwidth}
        \centering
        \includegraphics[width=\textwidth]{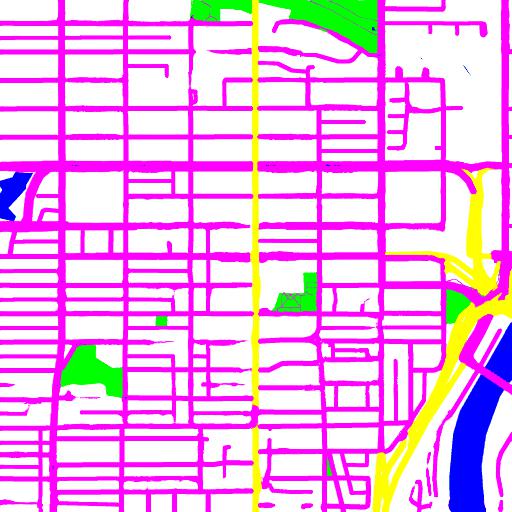}
    \end{subfigure}
    \begin{subfigure}[b]{0.23\textwidth}
        \centering
        \includegraphics[width=\textwidth]{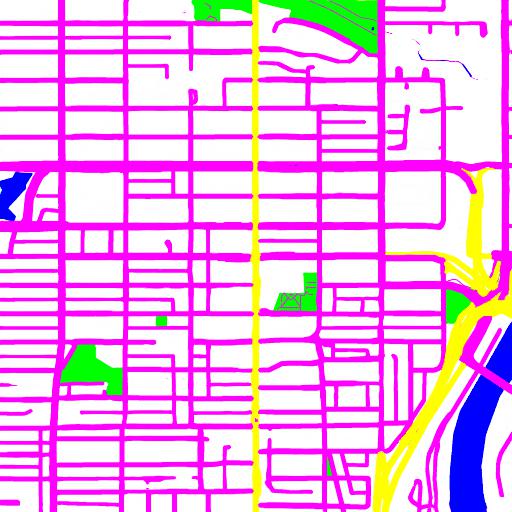}
    \end{subfigure}

    \vspace{4pt}

    \begin{subfigure}[b]{0.23\textwidth}
        \centering
        \includegraphics[width=\textwidth]{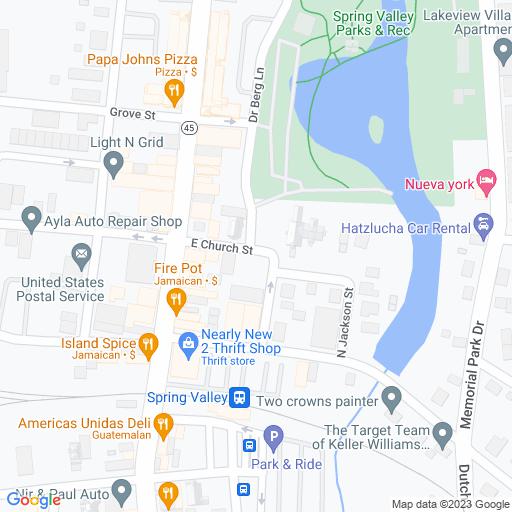}
    \end{subfigure}
    \begin{subfigure}[b]{0.23\textwidth}
        \centering
        \includegraphics[width=\textwidth]{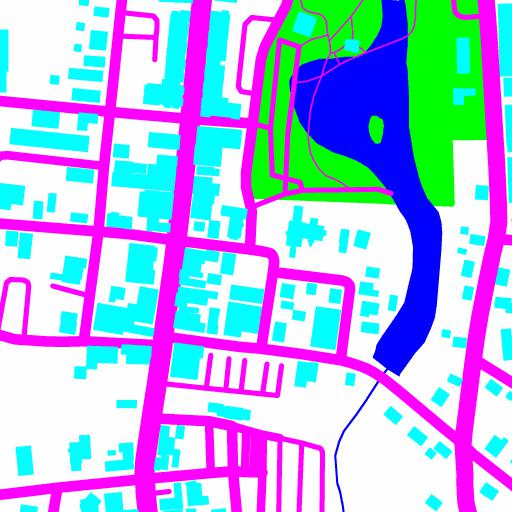}
    \end{subfigure}
    \begin{subfigure}[b]{0.23\textwidth}
        \centering
        \includegraphics[width=\textwidth]{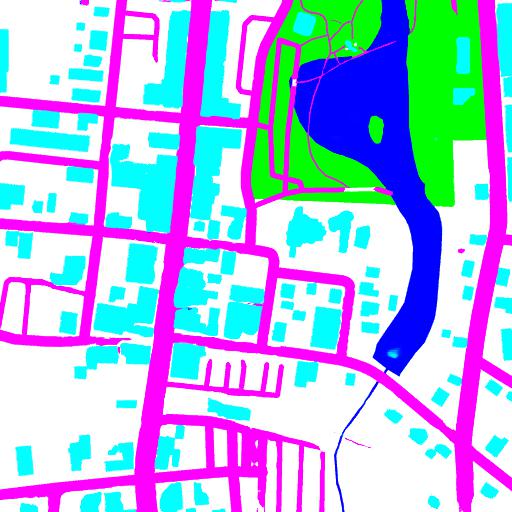}
    \end{subfigure}
    \begin{subfigure}[b]{0.23\textwidth}
        \centering
        \includegraphics[width=\textwidth]{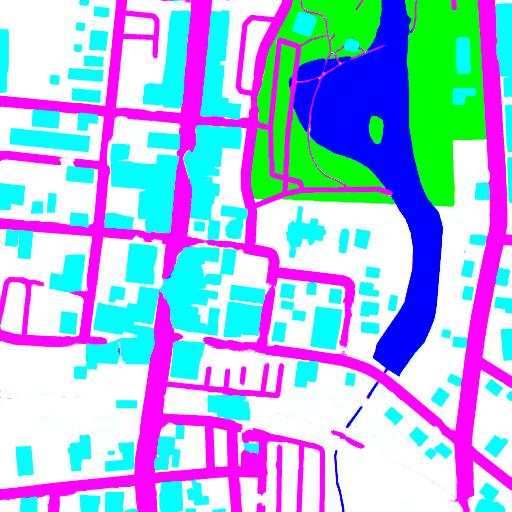}
    \end{subfigure}

    \vspace{4pt}

    \begin{subfigure}[b]{0.23\textwidth}
        \centering
        \includegraphics[width=\textwidth]{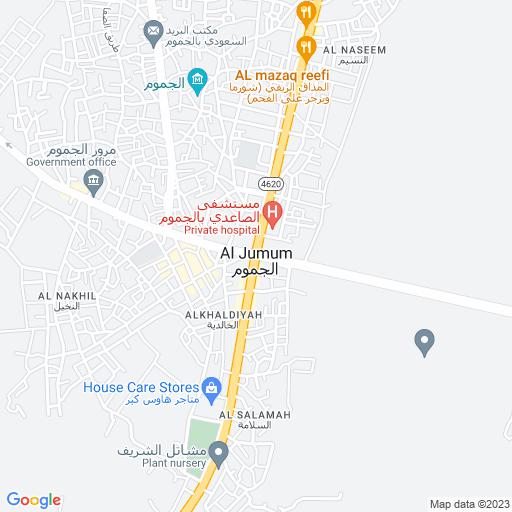}
    \end{subfigure}
    \begin{subfigure}[b]{0.23\textwidth}
        \centering
        \includegraphics[width=\textwidth]{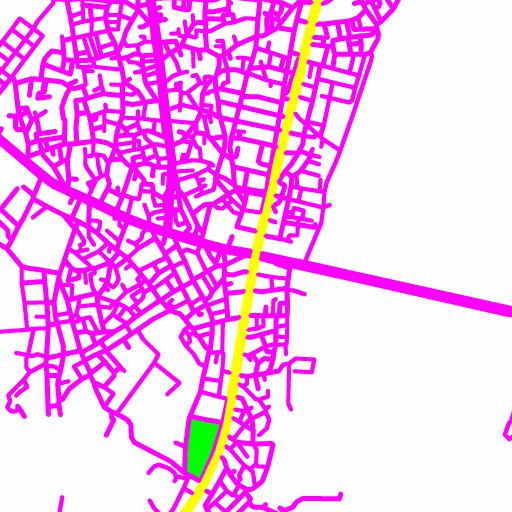}
    \end{subfigure}
    \begin{subfigure}[b]{0.23\textwidth}
        \centering
        \includegraphics[width=\textwidth]{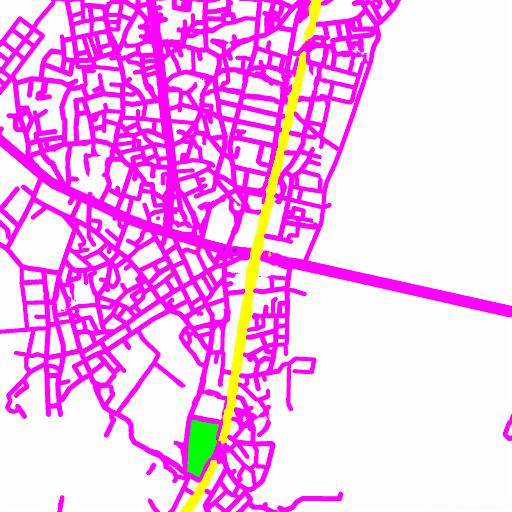}
    \end{subfigure}
    \begin{subfigure}[b]{0.23\textwidth}
        \centering
        \includegraphics[width=\textwidth]{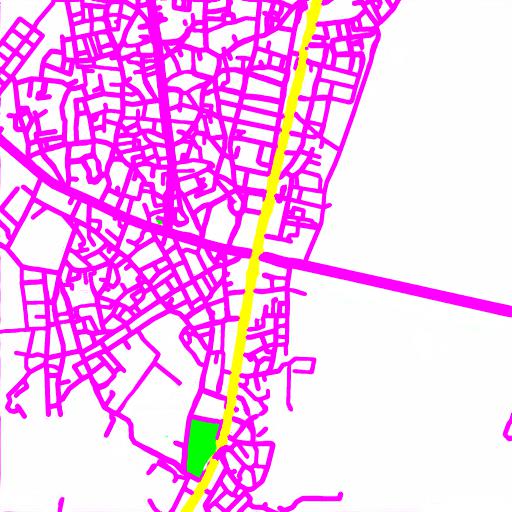}
    \end{subfigure}

    \vspace{4pt}

    \begin{subfigure}[b]{0.23\textwidth}
        \centering
        \includegraphics[width=\textwidth]{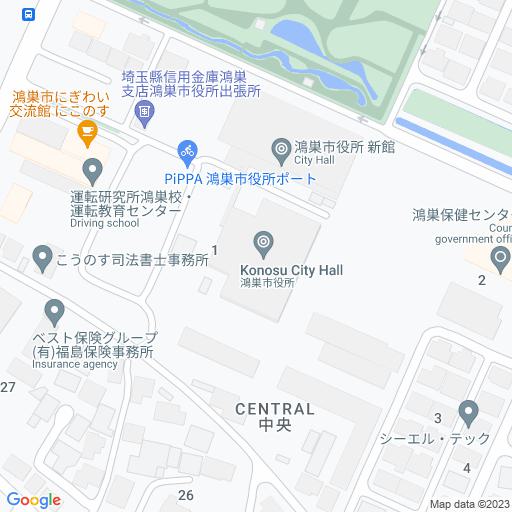}
    \end{subfigure}
    \begin{subfigure}[b]{0.23\textwidth}
        \centering
        \includegraphics[width=\textwidth]{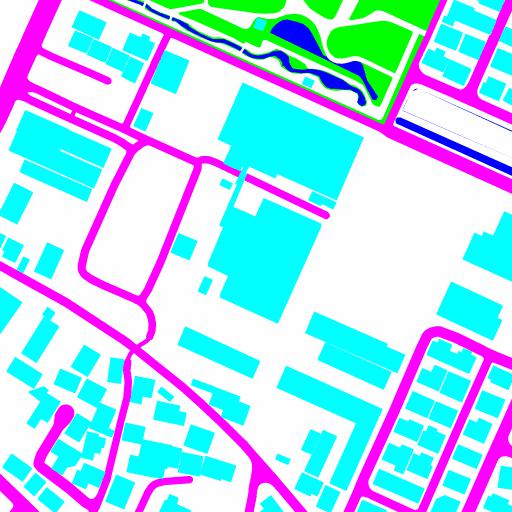}
    \end{subfigure}
    \begin{subfigure}[b]{0.23\textwidth}
        \centering
        \includegraphics[width=\textwidth]{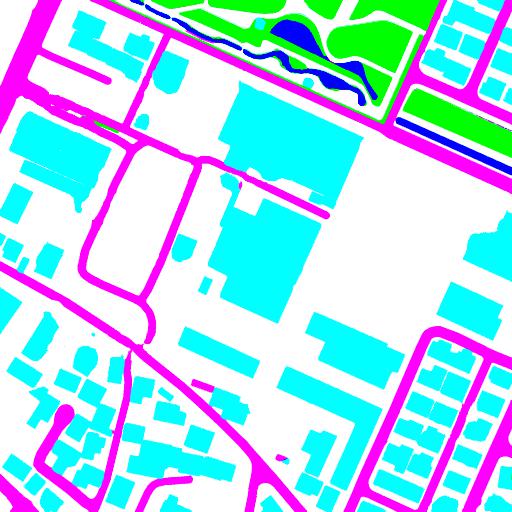}
    \end{subfigure}
    \begin{subfigure}[b]{0.23\textwidth}
        \centering
        \includegraphics[width=\textwidth]{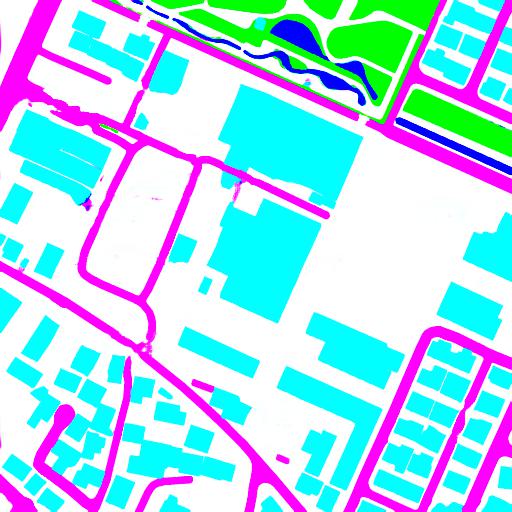}
    \end{subfigure}

    \vspace{4pt}

    \begin{subfigure}[b]{0.23\textwidth}
        \centering
        \includegraphics[width=\textwidth]{Filler_horizontal.jpg}
        \caption{Source Map}
        \label{fig:results_src2}
    \end{subfigure}
    \begin{subfigure}[b]{0.23\textwidth}
        \centering
        \includegraphics[width=\textwidth]{Google_copyright_logo.jpg}
        \caption{Ground Truth}
        \label{fig:results_gt2}
    \end{subfigure}
    \begin{subfigure}[b]{0.23\textwidth}
        \centering
        \includegraphics[width=\textwidth]{Filler_horizontal.jpg}
        \caption{Double Zoom Model}
        \label{fig:results_double2}
    \end{subfigure}
    \begin{subfigure}[b]{0.23\textwidth}
        \centering
        \includegraphics[width=\textwidth]{Filler_horizontal.jpg}
        \caption{Single Zoom Model}
        \label{fig:results_single2}
    \end{subfigure}
    \caption{Sample model outputs from the zoom 15 and 17 \textit{English} test sets, and the World test sets. The copyright underneath the second column applies to all images in that column.}
    \label{fig:good_15_17_ne}
\end{figure}

Tables \ref{tab:zoom_15_results}, \ref{tab:zoom_17_results}, and \ref{tab:zoom_17_ne_results} show the metrics when applied to the zoom 15, 17 and \textit{World} zoom 17 test sets, respectively. While not shown here, we include the metrics tables for the zoom 15, 16, and 18 \textit{World} test sets in \ref{sec:appendix_additional_results}. Figure \ref{fig:good_15_17_ne} further shows some sample outputs in these instances with rows 1 and 2 taken from the zoom 15 and 17 test sets, and rows 3 and 4 taken from the \textit{World} test sets (taken at zooms 16 and 18).

\begin{table}[t!]
\centering
\resizebox{\textwidth}{!}{
    \begin{tabular}{cc||ccc|ccc|ccc|ccc}
    \toprule
    & \textbf{Metric} & \multicolumn{3}{c|}{IoU} & \multicolumn{3}{c|}{F1} & \multicolumn{3}{c|}{Precision} & \multicolumn{3}{c}{Recall} \\
    & \textbf{Model} & Double & Single & \emph{Diff} & Double & Single & \emph{Diff} & Double & Single & \emph{Diff} & Double & Single & \emph{Diff} \\
    \hline
    \textbf{Class} & \textbf{Statistic} &  &  &  &  &  &  &  &  &  &  &  &  \\
    \midrule
    \multirow[c]{2}{*}{Streets} & median & 92.5 & 88.6 & 3.9 & 96.1 & 94.0 & 2.1 & 96.3 & 95.9 & 0.4 & 96.5 & 92.2 & 4.3 \\    
     & mean & 90.1 & 86.0 & 4.1 & 94.4 & 92.0 & 2.4 & 95.0 & 94.7 & 0.3 & 94.0 & 89.8 & 4.2 \\
    \multirow[c]{2}{*}{Highways} & median & 97.0 & 94.5 & 2.5 & 98.5 & 97.2 & 1.3 & 97.9 & 96.0 & 1.9 & 99.3 & 99.0 & 0.3 \\    
     & mean & 94.1 & 92.5 & 1.6 & 96.2 & 95.4 & 0.8 & 96.4 & 94.6 & 1.8 & 96.4 & 96.4 & 0.0 \\
    \multirow[c]{2}{*}{Parks} & median & 95.4 & 94.5 & 0.9 & 97.6 & 97.2 & 0.4 & 98.1 & 97.6 & 0.5 & 97.7 & 97.7 & 0.0 \\    
     & mean & 92.5 & 91.3 & 1.2 & 95.5 & 94.8 & 0.7 & 96.1 & 94.9 & 1.2 & 95.0 & 94.8 & 0.2 \\
    \multirow[c]{2}{*}{Water} & median & 99.0 & 98.4 & 0.6 & 99.5 & 99.2 & 0.3 & 99.4 & 99.2 & 0.2 & 99.7 & 99.4 & 0.3 \\    
     & mean & 98.0 & 96.2 & 1.8 & 99.0 & 97.9 & 1.1 & 98.6 & 98.2 & 0.4 & 99.4 & 97.8 & 1.6 \\
    \multirow[c]{2}{*}{Buildings} & median & 92.0 & 91.2 & 0.8 & 95.9 & 95.4 & 0.5 & 94.8 & 92.9 & 1.9 & 97.7 & 98.3 & -0.6 \\    
     & mean & 90.6 & 89.1 & 1.5 & 94.9 & 94.0 & 0.9 & 93.3 & 90.8 & 2.5 & 97.0 & 97.9 & -0.9 \\
    \multirow[c]{2}{*}{Hospitals} & median & 95.8 & 94.7 & 1.1 & 97.9 & 97.3 & 0.6 & 98.6 & 98.3 & 0.3 & 97.1 & 96.7 & 0.4 \\    
     & mean & 92.6 & 90.5 & 2.1 & 95.6 & 94.1 & 1.5 & 97.1 & 96.3 & 0.8 & 94.3 & 92.6 & 1.7 \\
     \bottomrule
    \end{tabular}
}
\caption{Class performances on the \textit{English} zoom-17 test set.}
\label{tab:zoom_17_results}
\end{table}

\begin{table}
\centering
\resizebox{\textwidth}{!}{
    \begin{tabular}{cc||ccc|ccc|ccc|ccc}
    \toprule
    & \textbf{Metric} & \multicolumn{3}{c|}{IoU} & \multicolumn{3}{c|}{F1} & \multicolumn{3}{c|}{Precision} & \multicolumn{3}{c}{Recall} \\
    & \textbf{Model} & Double & Single & \emph{Diff} & Double & Single & \emph{Diff} & Double & Single & \emph{Diff} & Double & Single & \emph{Diff} \\
    \hline
    \textbf{Class} & \textbf{Statistic} &  &  &  &  &  &  &  &  &  &  &  &  \\
    \midrule
    \multirow[c]{2}{*}{Streets} & median & 84.8 & 77.4 & 7.4 & 91.8 & 87.2 & 4.6 & 91.1 & 94.1 & -3.0 & 96.0 & 82.3 & 13.7 \\   
     & mean & 82.5 & 74.1 & 8.4 & 89.9 & 84.1 & 5.8 & 87.2 & 91.8 & -4.6 & 93.8 & 78.7 & 15.1 \\
    \multirow[c]{2}{*}{Highways} & median & 95.1 & 91.3 & 3.8 & 97.5 & 95.5 & 2.0 & 96.8 & 95.8 & 1.0 & 99.0 & 96.7 & 2.3 \\   
     & mean & 93.1 & 86.7 & 6.4 & 96.2 & 92.3 & 3.9 & 95.4 & 94.7 & 0.7 & 97.4 & 91.2 & 6.2 \\
    \multirow[c]{2}{*}{Parks} & median & 96.0 & 94.5 & 1.5 & 98.0 & 97.2 & 0.8 & 98.2 & 97.5 & 0.7 & 98.1 & 98.0 & 0.1 \\   
     & mean & 92.6 & 89.5 & 3.1 & 95.3 & 93.3 & 2.0 & 95.8 & 94.4 & 1.4 & 95.3 & 93.3 & 2.0 \\
    \multirow[c]{2}{*}{Water} & median & 97.4 & 96.3 & 1.1 & 98.7 & 98.1 & 0.6 & 98.3 & 98.5 & -0.2 & 99.5 & 99.0 & 0.5 \\   
     & mean & 95.0 & 91.9 & 3.1 & 97.3 & 95.3 & 2.0 & 96.4 & 96.0 & 0.4 & 98.5 & 95.4 & 3.1 \\
    \multirow[c]{2}{*}{Buildings} & median & 75.3 & 78.5 & -3.2 & 85.9 & 87.9 & -2.0 & 93.4 & 84.3 & 9.1 & 84.5 & 94.1 & -9.6 \\   
     & mean & 64.2 & 76.3 & -12.1 & 73.0 & 85.7 & -12.7 & 90.0 & 81.2 & 8.8 & 69.6 & 92.2 & -22.6 \\
    \multirow[c]{2}{*}{Hospitals} & median & 89.5 & 76.7 & 12.8 & 94.5 & 86.8 & 7.7 & 94.3 & 97.8 & -3.5 & 96.5 & 77.4 & 19.1 \\   
     & mean & 83.0 & 62.3 & 20.7 & 88.9 & 69.3 & 19.6 & 91.7 & 91.1 & 0.6 & 89.1 & 63.6 & 25.5 \\
     \bottomrule
    \end{tabular}
}
\caption{Class performances on the \textit{World} zoom 17 test set.}
\label{tab:zoom_17_ne_results}
\end{table}

As with results for zooms 16 and 18, both single and double zoom models perform well at zooms 15 and 17 albeit with small drops in performance. Compared to at zoom 16, both models experience drops across all metrics on the zoom 15 test set, usually by about 3-7\% in IoU. A similar trend appears in the zoom 17 metrics compared to the zoom 18 metrics. Like in the previous case, the Zoom-16/18 model slightly underperforms the Zoom-16 model at zoom 15, but outperforms the Zoom-18 model on zoom 17 maps. Note that for the hospital class in Table \ref{tab:zoom_15_results}, the low mean is likely a result of errors in the color segmentation scheme.

For the zoom 15 results, the largest drops occurred specifically in the IoU and precision compared to at zoom 16, particularly for streets and highways. This is likely linked to a greater number of mistakes related to inpainting over street names. 
Conversely at zoom 17 compared to zoom 18, we observe equal drops in precision and recall, particularly for the street and buildings classes. This was likely a result of the model mistaking these two classes, since both can appear as grey in the source maps. Overall, the double zoom model handles grey streets more effectively than the Zoom-18 model, which is attributable to data augmentation.

Overall, in terms of the general performance drops at zooms 15 and 17, we hypothesize that these occur due to a higher density of map features at these zooms compared to zooms 16 and 18. This is inherently due to the fact that zooms 15 and 17 provide a more zoomed out perspective with more fine details, thus making inpainting more challenging. Given these problems, we expect that training involving mixtures of these zooms would remedy the problems and improve the metrics. 

\subsection{Generalizing to other World Regions and Languages}

The last two rows of Figure \ref{fig:good_15_17_ne} illustrate two examples from the \textit{World} test sets, specifically where the main languages are not English. As can be seen, these models show no difficulty in identifying and removing these texts, and indeed, in general have no problems removing any text from any language observed, including those not based on the Latin script. 

\begin{figure}[t!]
    \centering 

    \begin{subfigure}[b]{0.23\textwidth}
        \centering
        \includegraphics[width=\textwidth]{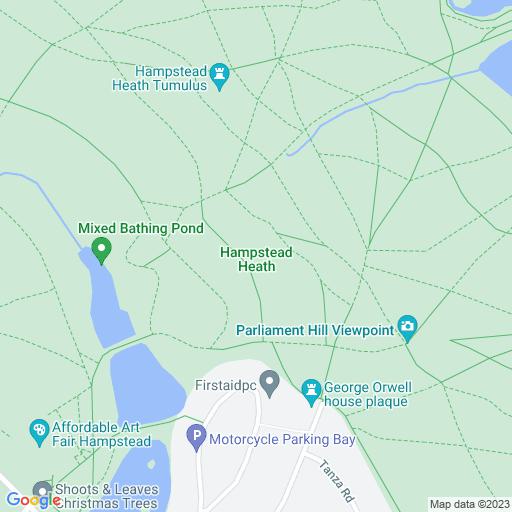}
    \end{subfigure}
    \begin{subfigure}[b]{0.23\textwidth}
        \centering
        \includegraphics[width=\textwidth]{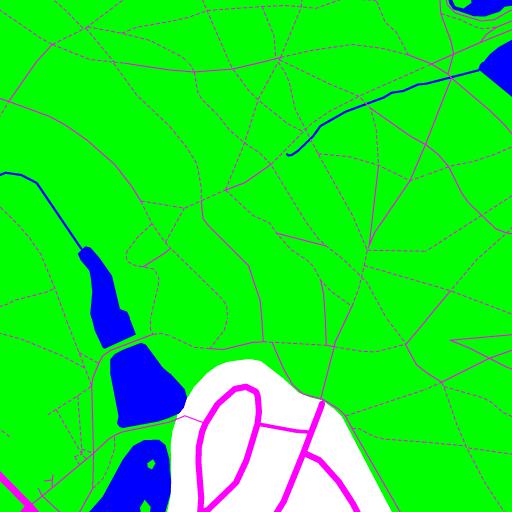}
    \end{subfigure}
    \begin{subfigure}[b]{0.23\textwidth}
        \centering
        \includegraphics[width=\textwidth]{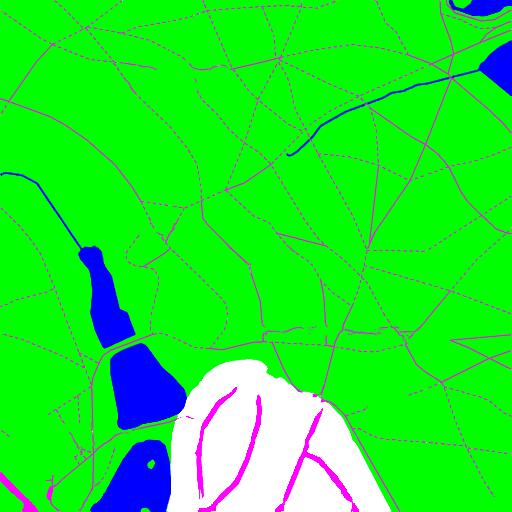}
    \end{subfigure}
    \begin{subfigure}[b]{0.23\textwidth}
        \centering
        \includegraphics[width=\textwidth]{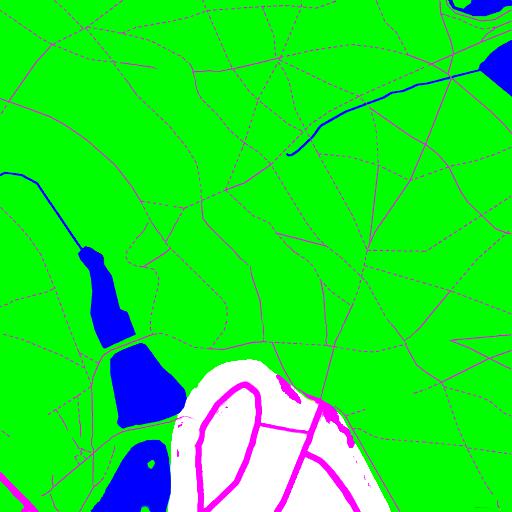}
    \end{subfigure}

    \vspace{4pt}

    \begin{subfigure}[b]{0.23\textwidth}
        \centering
        \includegraphics[width=\textwidth]{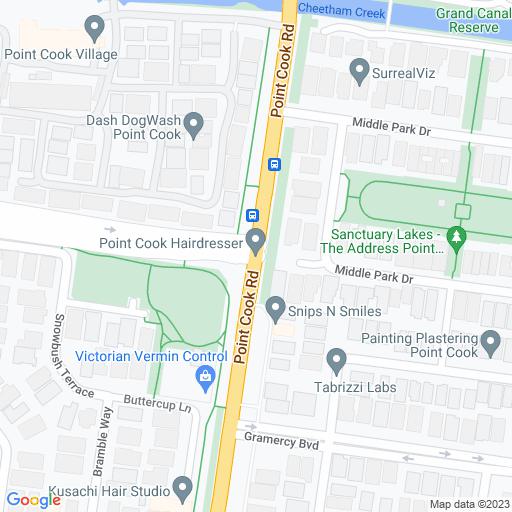}
    \end{subfigure}
    \begin{subfigure}[b]{0.23\textwidth}
        \centering
        \includegraphics[width=\textwidth]{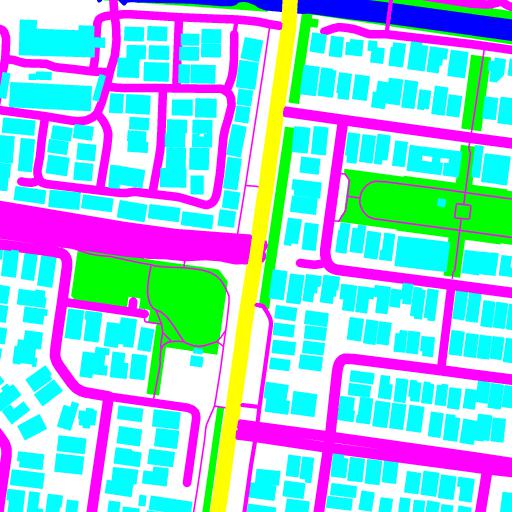}
    \end{subfigure}
    \begin{subfigure}[b]{0.23\textwidth}
        \centering
        \includegraphics[width=\textwidth]{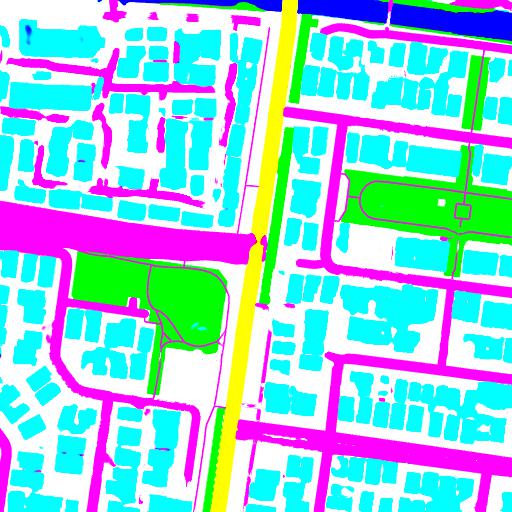}
    \end{subfigure}
    \begin{subfigure}[b]{0.23\textwidth}
        \centering
        \includegraphics[width=\textwidth]{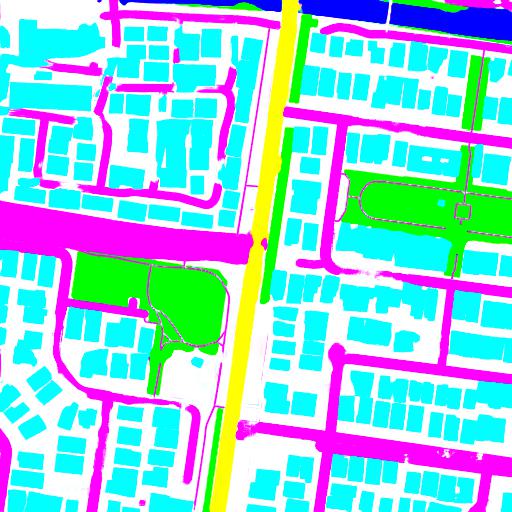}
    \end{subfigure}
    
    \vspace{4pt}

    \begin{subfigure}[b]{0.23\textwidth}
        \centering
        \includegraphics[width=\textwidth]{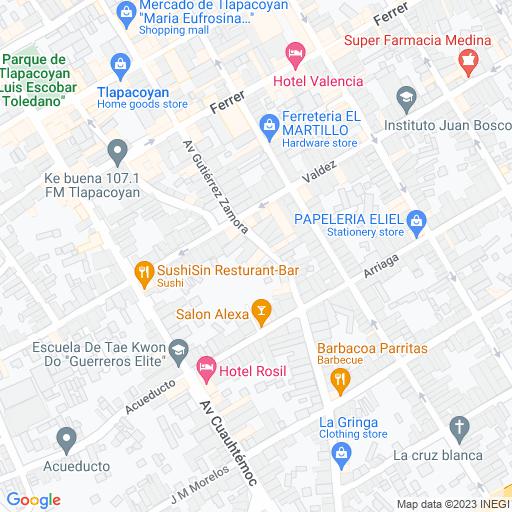}
    \end{subfigure}
    \begin{subfigure}[b]{0.23\textwidth}
        \centering
        \includegraphics[width=\textwidth]{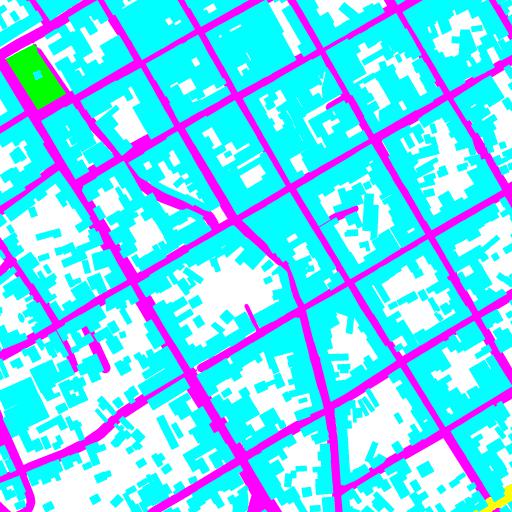}
    \end{subfigure}
    \begin{subfigure}[b]{0.23\textwidth}
        \centering
        \includegraphics[width=\textwidth]{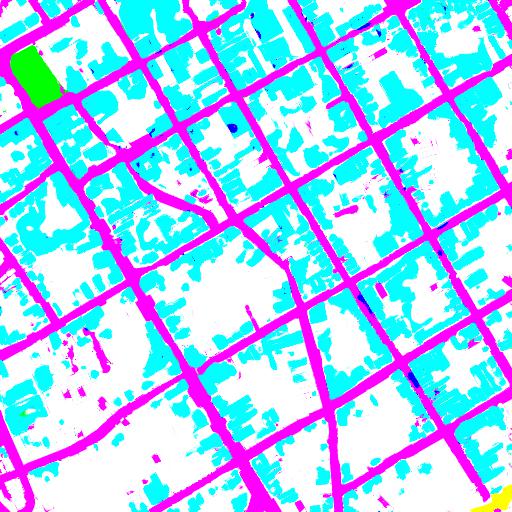}
    \end{subfigure}
    \begin{subfigure}[b]{0.23\textwidth}
        \centering
        \includegraphics[width=\textwidth]{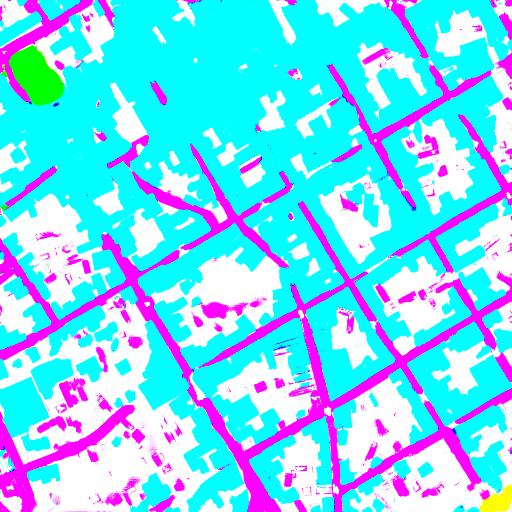}
    \end{subfigure}

    \vspace{4pt}

    \begin{subfigure}[b]{0.23\textwidth}
        \centering
        \includegraphics[width=\textwidth]{Filler_horizontal.jpg}
        \caption{Source Map}
        \label{fig:results_src3}
    \end{subfigure}
    \begin{subfigure}[b]{0.23\textwidth}
        \centering
        \includegraphics[width=\textwidth]{Google_copyright_logo.jpg}
        \caption{Ground Truth}
        \label{fig:results_gt3}
    \end{subfigure}
    \begin{subfigure}[b]{0.23\textwidth}
        \centering
        \includegraphics[width=\textwidth]{Filler_horizontal.jpg}
        \caption{Double Zoom Model}
        \label{fig:results_double3}
    \end{subfigure}
    \begin{subfigure}[b]{0.23\textwidth}
        \centering
        \includegraphics[width=\textwidth]{Filler_horizontal.jpg}
        \caption{Single Zoom Model}
        \label{fig:results_single3}
    \end{subfigure}  
    \caption{Sample model outputs illustrating limitations of the double and single zoom models. The copyright underneath the second column applies to all images in that column.}

    \label{fig:bad}
\end{figure}

The performance on these test sets is comparable to those from the \textit{English} test sets, overall, with only minor drops in performance. All IoU metrics are above 87\% and we hypothesize that these drops are similar to those at zooms 15 and 17 in that they can be attributed to a higher density of features in the maps. This is likely the case since population densities in the United States, Canada, and Australia are low compared to other countries. 

Due to the similarity with Tables \ref{tab:zoom_16_results}, \ref{tab:zoom_18_results}, and \ref{tab:zoom_15_results}, the tables for the \textit{World} test set metrics at zooms 16, 18, and 15 are not presented here and are instead given in \ref{sec:appendix_additional_results} for interested readers. Here, however, we show the metric table for the \textit{World} zoom 17 test set (Table \ref{tab:zoom_17_ne_results}) since these results highlight an additional difficulty related to distinguishing between buildings and streets. Notably, the median IoU metrics for streets are below 85\%, and for buildings are below 79\% for both models. The last row in Figure \ref{fig:bad} shows an example of these issues and demonstrates that the models struggle with maps containing small buildings, particularly in high density. While the Zoom-18 model struggles to identify streets properly, the Zoom-16/18 model better traces the streets but is less adept at locating the building boundaries. This is further complimented by the metrics in Table \ref{tab:zoom_17_ne_results} which show low recall for buildings in the double zoom model, and simultaneously low street recall and building precision for the Zoom-18 model. Again however, we hypothesize that this can be remedied by training on maps with higher densities of features.

\section{Limitations and Future Work} \label{sec:limitations}
While our results are promising and address many of the shortcomings we mention in the Related Works section, our work remains a proof-of-concept and provides only a first step towards developing a readily accessible tactile map generation tool.

Besides addressing the minor issues outlined in section \ref{sec:results}, incorporating additional zooms and expanding the number of features to point symbols (in addition to the line- and area-like features we include here) are important areas for future work. In general, consultation with the community of PVI and individual users will be essential to establish the most important and relevant features \citep{rowell_feeling_your_way}. 

As noted earlier, our models only process maps from Google. While one may note that the method used to generate the ground truth tactile maps is more accurate than any of our models, it is only applicable to Google Maps and not to any other source. The potential to address this concern, and train models capable of generalizing to completely new map styles like Bing Maps, is one of the major advantages of using CV-based techniques. This is contingent on the availability of quality data, and hence expanding our dataset to maps outside of Google Maps will be an important future endeavour.

Beyond these CV-related limitations, much work remains for a full realization of a tactile map solution. We need to develop and evaluate separate tasks and methods for identifying street names, creating legends, marking orientation arrows and scale lines, and making conversions of text to Braille. Research in optical character recognition \citep{text_extraction_review}, and other off-the-shelf methods, provide promising avenues for accomplishing many of these tasks\footnote{https://cloud.google.com/vision/docs/ocr \newline https://developer.apple.com/documentation/vision/recognizing\_text\_in\_images}. More research, however, is necessary to see how these approaches can handle the collection of tasks required for the tactile map use-case and which are most suitable.

While much work remains, we are motivated by the incredible potential these solutions offer, including being easy to apply, flexible to new environments, and providing opportunities for innovative and impactful new solutions in the field of accessibility. As an example, \citet{fluxmarker} presents dynamic tactile maps capable of updating their content in real-time. Such technologies could be incorporated into our method though an interface with Google. Not only would this create an interactive and adaptable solution, but it could also leverage the wealth of other information Google can provide, including store hours, customer reviews, and other general information of interest.

\section{Conclusion}
In conclusion, much work remains to address existing challenges related to accessibility, especially for people with visual impairments and in the tactile mapping space more specifically.

In a step towards addressing these concerns, we explore the use of Pix2Pix-GAN models to help automate the generation of tactile maps from visual maps at the street level. To this end, we create a large-scale dataset of source-tactile map pairs from Google Maps consisting of 6500 different locations over four zoom scales and six line- and area-like tactile features. We show that our models are proficient at tasks necessary for automated tactile map generation, particularly those that are applicable across map guidelines. At the aggregated tasks of recognizing important features, removing extraneous ones, and simplifying the resulting maps, models trained at two zoom levels achieve IoU and F1 scores of better than 0.95 across all classes with only slight drops in performance compared to at a single zoom. We further show the models are adept at interpolating and extrapolating to unseen scales and generalizing to maps containing non-English text. 

While we require further efforts to develop a complete automated tactile map solution using CV techniques, the work here serves as a foundation for future research, hopefully paving the way for more advanced solutions that will benefit both the academic and industrial communities focused on accessibility. Future efforts will involve expanding the dataset to maps outside of Google, and including the recognition of point symbols.

Ultimately, the goal of these models is to one day provide generalizability beyond just mapping, and to develop a general tactile solution capable of creating tactile graphics across multiple domains such as building interiors, maps, tactile graphics, and more. Current avenues for extending these results include exploring exciting new developments in machine learning, such as StarGAN \citep{stargan} for multi-domain image-to-image translation, or other CV architectures like diffusion models \citep{diffusion}.

\section{Acknowledgements}
This work was supported by the Natural Sciences and Engineering Research Council of Canada (NSERC).

\bibliographystyle{elsarticle-harv} 
\bibliography{references}

\newpage
\appendix
\section{Dataset}

The following subsection provides a discussion of the tactile map features and standards used in prior works.  

\subsection{Data Collection Details using the Google Maps Static API} \label{sec:appendix_api}
Table \ref{tab:style-table} gives the complete style specification used for retrieving the tactile maps using the Google Maps Static API. These specifications are passed using the \texttt{style} parameter of the API requests. For additional information on using this style specification, see the link in the footnote\footnote{https://developers.google.com/maps/documentation/maps-static/styling}.

\begin{table}[h!]
\centering
\begin{tabular}{ccc}
\toprule
\textbf{Feature} & \textbf{Element} & \textbf{Specification} \\
\midrule
all      &     all      &     color:0xffffff      \\ 
administrative       &     labels      &    visibility:off       \\ 
landscape       &     all      &     color:0x000000      \\ 
landscape      &     labels      &    visibility:off       \\ 
landscape.man\_made       &     all      &   color:0x00ffff        \\ 
landscape.man\_made       &     geometry.fill      &    color:0xffffff  \\ 
landscape.man\_made.building       &     geometry.fill      &    color:0x00ffff \\ 
landscape.natural      &     all      &   color:0xffffff        \\ 
poi      &      labels     &     visibility:off       \\ 
poi     &    geometry.fill       &     color:0x00ff00      \\ 
poi.medical      &    geometry.fill       &     color:0x808080      \\ 
poi.place\_of\_worship      &    geometry.fill       &     visibility:off      \\ 
poi.school      &    geometry.fill       &     visibility:off      \\ 
road      &     labels      &     visibility:off       \\ 
road.highway      &     all      &      color:0xffff00     \\ 
road.highway      &     geometry.fill      &     color:0xffff00       \\ 
road.highway.controlled\_access       &     geometry.fill      &     color:0xffff00       \\ 
road.arterial      &     all      &      color:0xff00ff      \\ 
road.arterial      &     geometry.fill      &      color:0xff00ff      \\ 
road.local   &     all      &     color:0xff00ff       \\ 
road.local      &      geometry.fill     &     color:0xff00ff       \\ 
transit      &     all      &     visibility:off       \\ 
transit       &    labels       &    visibility:off       \\ 
water      &      all     &    color:0x0000ff       \\ 
water      &     geometry.fill      &     color:0x0000ff      \\
water      &     labels      &    visibility:off       \\ 
\bottomrule
\end{tabular}
\caption[Google Maps Static API Style Specification for Tactile Maps]{Complete style specification for calls to the Google Maps Static API to create the tactile maps.}
\label{tab:style-table}
\end{table}

Landmarks were retrieved from the following cities in each country:

\begin{itemize}
    \item Australia: Adelaide, Brisbane, Cairns, Canberra, Darwin, Hobart, Melbourne, Perth, Sydney
    \item Canada: Calgary, Montréal, Toronto, Ottawa, Vancouver
    \item UK: Edinburgh, Leicester, Birmingham, Glasgow, London, Manchester, Bristol, Liverpool
    \item US: Chicago, Los Angeles, Las Vegas, Seattle, New York, Columbus, Washington D.C., Nashville, San Diego, San Francisco, Boston, Michigan (state), Hawaii (state), Florida (state), Indiana (state)
\end{itemize}

Queries to the Google API were formatted in the following way:

\begin{itemize}
    \item Cities: 
        \begin{itemize}
            \item UK:``\emph{$<$City$>$, $<$Country$>$, UK}"
            \item All other countries: ``\emph{$<$City$>$, $<$Province/State$>$, $<$Country$>$}"
        \end{itemize}
    \item Landmarks: ``\emph{$<$Landmark Name$>$, $<$City/State$>$, $<$Country$>$}"
    \item Universities/Colleges: ``\emph{$<$Institution Name$>$, $<$Country$>$}"
    \item Hospitals: ``\emph{$<$Hospital Name$>$, $<$Address$>$, $<$City$>$, $<$State$>$, USA}"
\end{itemize}

\subsection{Location Details} \label{sec:appendix_loc_details}
We chose locations for the maps to include a broad range of environments while also ensuring an adequate representation for each feature in the dataset. Locations are a combination of cities, landmarks, hospitals, and universities and colleges; all of which being areas in which PVI may enjoy greater access \citep{accessibility, papadopoulos}. For the training and \textit{English} test sets, cities, landmarks, and universities and colleges are taken from all countries, however hospitals are only taken from US locations since only data from the US was readily accessible.

We gather city names from all countries from an open world cities dataset provided by Alexandre Bonnasseau and the Open Knowledge Foundation\footnote{https://github.com/datasets/world-cities}.
This dataset consists of a list of all world cities with populations above 15,000, as collected by the GeoNames geographical database. 
For landmarks, we collect them manually from different travel websites and blogs listing famous landmarks in major cities or states for travel purposes\footnote{https://www.destguides.com/en \newline https://nomadsunveiled.com/ \newline https://www.kevmrc.com/}. Universities and colleges are taken from a public dataset of world universities which had been compiled from multiple sources including Wikipedia's List of Universities in the World\footnote{https://www.kaggle.com/datasets/thedevastator/all-universities-in-the-world}. 
Finally for the hospitals, we take locations from the USA Hospitals dataset provided by the Homeland Infrastructure Foundation-Level Data website\footnote{https://www.kaggle.com/datasets/carlosaguayo/usa-hospitals}. 

The \textit{World} test set is collected in the same way, however only using city locations randomly selected from the open world cities dataset after filtering out cities from the countries in the training set. This is also done at zooms 15, 16, 17, and 18. 

\subsection{Dataset Breakdowns} \label{sec:appendix_dataset_breakdown}

Table \ref{tab:dataset_location_breakdown} gives the breakdown of locations in the training and \textit{English} test sets by country and by type.

Table \ref{tab:dataset_class_breakdown} displays the class breakdowns for the training sets at zooms 16 and 18. Specifically, the values represent the average percentage of pixels across all tactile maps in the training set that correspond to that class. Since models are never trained on zooms 15 and 17, only the breakdowns for these zooms are provided.

Table \ref{tab:world_test_set_counts} gives the top 10 countries with the most locations in the \textit{World} test set.

\begin{table}[ht!]
\centering
\begin{tabular}{ccc}
    \toprule
    \textbf{Location Type} & \textbf{Country} &   \textbf{Count}    \\
    \midrule
    Cities & Australia &   174 \\
                              & Canada &   172 \\
                              & UK &   371 \\
                              & USA &  1744 \\
    \hline
    Hospitals & USA &  1543 \\
    \hline
    Landmarks & Australia &   116 \\
                              & Canada &    91 \\
                              & UK &   168 \\
                              & USA &   280 \\
    \hline
    Universities and Colleges & Australia &    29 \\
                              & Canada &    96 \\
                              & UK &   132 \\
                              & USA &   584 \\
    \bottomrule
\end{tabular}
\caption[Dataset Breakdown of Locations]{Breakdown of the training and \textit{English} test sets by number of cities, landmarks, universities/colleges, and hospitals from each respective country. This applies to both the training and test sets together.}
\label{tab:dataset_location_breakdown}
\end{table}

\begin{table}[hb!]
\centering
\begin{tabular}{ccc}
\toprule
\textbf{Class}    & \multicolumn{2}{c}{\textbf{Mean Pixel Percentage (\%)}}    \\
 & Zoom 16 Dataset & Zoom 18 Dataset\\
 \midrule
Streets & 16.64 & 12.44 \\ 
Highways & 1.95 & 1.39 \\ 
Parks & 5.92 & 6.46 \\ 
Water & 3.71 &  1.74 \\ 
Buildings & N/A &  27.04 \\ 
Hospitals & 3.99 &  10.08 \\ 
Background & 67.79 &  40.85 \\
\bottomrule
\end{tabular}
\caption[Class Breakdown in the Training Set]{Class breakdown in the training sets at zoom 16 and zoom 18.}
\label{tab:dataset_class_breakdown}
\end{table}

\clearpage

\begin{table}[ht!]
\centering
\begin{tabular}{cc}
\toprule
\textbf{Country} & \textbf{Count} \\
\midrule
India     &   131 \\
Brazil    &    69 \\
Germany   &    57 \\
Japan     &    42 \\
Russia    &    40 \\
China     &    35 \\
Mexico    &    34 \\
Spain     &    33 \\
Indonesia &    27 \\
Italy     &    22 \\
\bottomrule
\end{tabular}
\caption{Top 10 countries with the most locations in the \textit{World} test set.}
\label{tab:world_test_set_counts}
\end{table}

\subsection{Further Discussion of User Requirements, Standards and Map Features for Tactile Maps} \label{sec:appendix_further_discussion}

In designing and selecting features within tactile maps, understanding the needs of users and PVI is of principle importance. 
In their work, \citet{stampach} notes that the location of buildings and main transportation networks are of highest importance to PVI in a university setting. \citet{rowell_feeling_your_way} find that general tactile map users consider city maps relevant to their needs but want more tactile maps for large outdoor spaces in addition to street-views. Of concern to these latter participants was also the changing nature of urban environments and its resulting affects on their mobility \citep{rowell_feeling_your_way}. Other works identify that the range of environments mapped for tactile production can be very broad and cover rural settings with non-trivial frequency \citep{rowell}. 

These concerns, coupled with their breadth and complexity, serve as a partial inspiration for our choice of Google Maps for the source maps in this study. Google provides not only large and accurate coverage of urban and rural spaces, including all the elements listed above, but also includes updated information as changes arise. This adaptability is an appealing feature for the use of Google Maps, as seen in a study of participants with visual impairments \citep{fluxmarker}.

In selecting the features of the tactile maps, many different sets of features and standards are used in prior work. TMAP, TMACS, Mapy.cz, and the tool by \citet{stampach} all use different features for area textures and line or point symbols \citep{mapy, tmap, tmacs}. They also all adhere to different tactile standards, including guidelines from Sweden \citep{eriksson}, Australia \citep{australia_guideline}, Japan \citep{japan_guideline}, the Braille Authority of North America \citep{bana}, as well as local support centers, and results of independent studies \citep{ishibashi2013easy, watanabe2012development}. This is further complicated by the fact that many other countries have their own guidelines, and international standards are neither globally recognized nor universally applicable \citep{wabinski}.

As noted by \citet{wabinski}, while many modern technological approaches to creating tactile maps may be innovative, they often do not address the full issues related to tactile map creation. Our work recognizes this fact, and while we do not to perform any consultations with PVI at this stage, our aim is to provide a proof of concept and show the proficiency of the candidate CV architecture. We acknowledge the need for further validation of our approach, in consultation with real users, for the production of fully usable tactile maps. 

\subsection{Map Legibility} \label{sec:appendix_map_legibility}
At this time, we do not complete any analyses to assess the overall quality and legibility of the generated tactile map images. This is because no automated tools capable of assessing the thousands of generated images are available for doing this. However we acknowledge that some of the tactile maps in the dataset, and presented here, may not be legible in a tactile modality without additional processing.

For the purposes of this paper, however, the scope is limited to assessing the effectiveness of the CV models at understanding and distinguishing different map regions. As the models trained here are general enough to handle maps regardless of legibility, the model performances reported below apply equally well to legible maps.

In physical realizations of this approach, efforts towards ensuring that tactile map datasets are legible, of high quality, and meaningful to users with visual impairments will be of paramount importance. As tactile map legibility also depends on the method of rendering, a legibility analysis would also be required at that stage based on the specific rendering method.

\section{Training Details} \label{sec:appendix_training_details}
We use basic data augmentation involving horizontal rotations, shifts, scale changes and rotations up to 15 degrees for the training of all models. For the Zoom-16/18 model, we apply additional augmentation to change the colors of streets and buildings both to grey. We do this on half the input maps at zoom 18 to help the model distinguish these classes by shape. Note that we leave the other classes unchanged by this augmentation since the identification of parks, water, and hospitals is not always possible purely by shape.

For the Zoom-16 and Zoom-18 models, we train on the training set corresponding to that zoom (5000 source-tactile images). For the Zoom-16/18 model, however, we train on the combination of the training sets from the two different zooms together (10,000 source-tactile images). 

Models are trained for 125 epochs with a batch size of 1, for easier memory management, and the Adam optimizer \citep{adam}. All hyperparameters for the optimizers and model architecture are their default values.

All code is written in Python with the neural network architectures implemented in PyTorch. The computer hardware made use of Nvidia RTX 3060 Ti GPUs leveraging the Cuda Toolkit. With these optimizations, training on a single zoom took between 3-4 days, whereas training on double zooms took 5-6 days.

\section{Additional Results from the \textit{World} test set} \label{sec:appendix_additional_results}

Tables \ref{tab:zoom_15_ne_results}, \ref{tab:zoom_16_ne_results}, and \ref{tab:zoom_18_ne_results} give the metric tables from the \textit{World} test sets at zooms 15, 16, and 18, respectively.

\begin{figure}[h!]
    \centering 

    \begin{subfigure}[b]{0.23\textwidth}
        \centering
        \includegraphics[width=\textwidth]{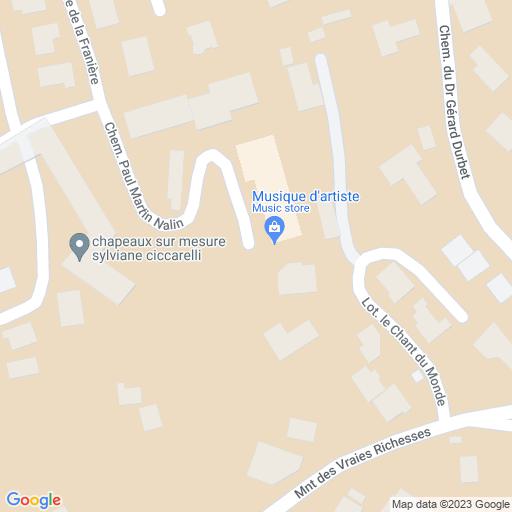}
    \end{subfigure}
    \begin{subfigure}[b]{0.23\textwidth}
        \centering
        \includegraphics[width=\textwidth]{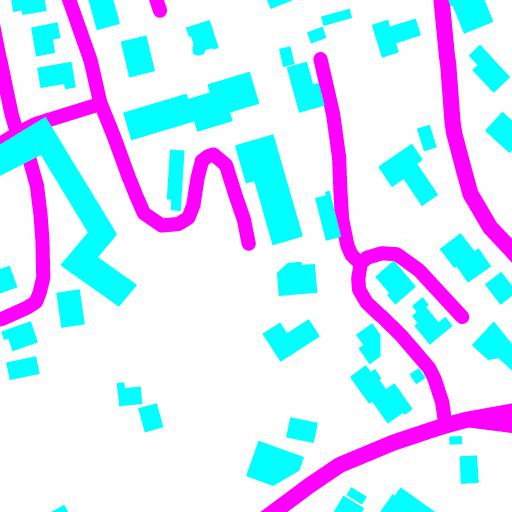}
    \end{subfigure}
    \begin{subfigure}[b]{0.23\textwidth}
        \centering
        \includegraphics[width=\textwidth]{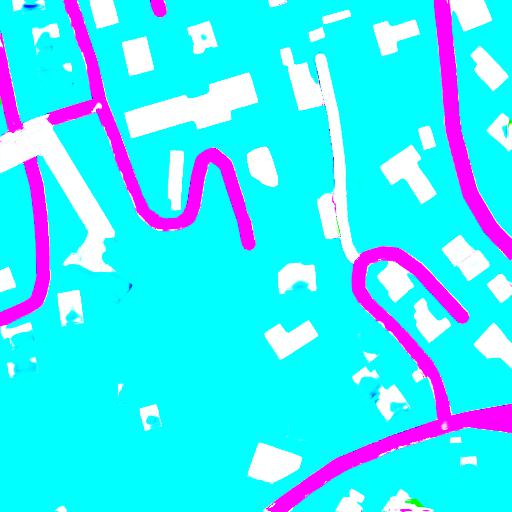}
    \end{subfigure}
    \begin{subfigure}[b]{0.23\textwidth}
        \centering
        \includegraphics[width=\textwidth]{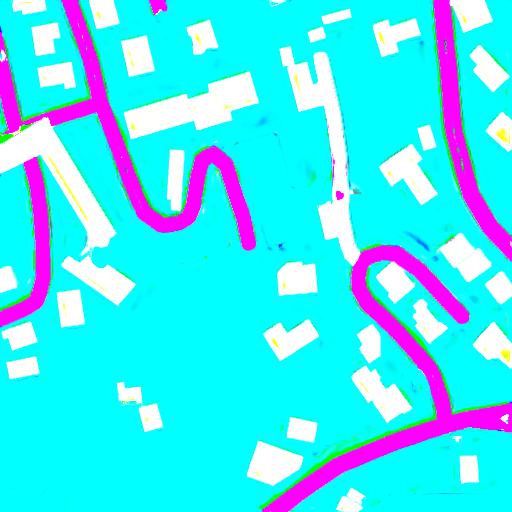}
    \end{subfigure}

    \vspace{4pt}

    \begin{subfigure}[b]{0.23\textwidth}
        \centering
        \includegraphics[width=\textwidth]{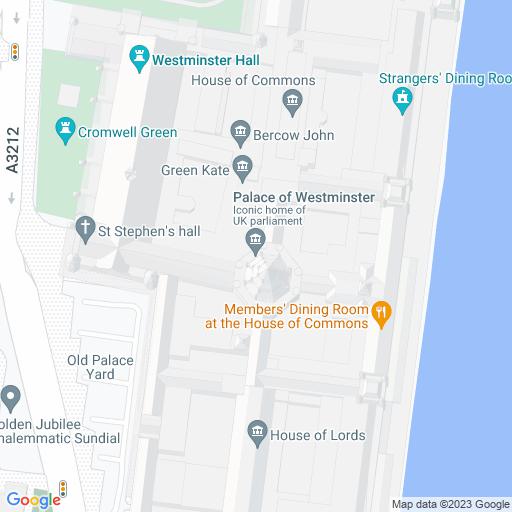}
    \end{subfigure}
    \begin{subfigure}[b]{0.23\textwidth}
        \centering
        \includegraphics[width=\textwidth]{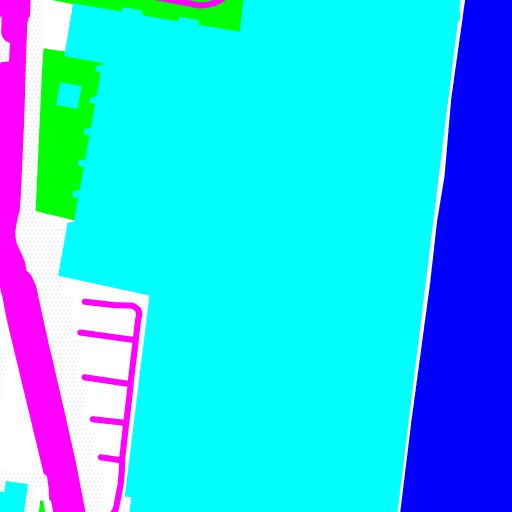}
    \end{subfigure}
    \begin{subfigure}[b]{0.23\textwidth}
        \centering
        \includegraphics[width=\textwidth]{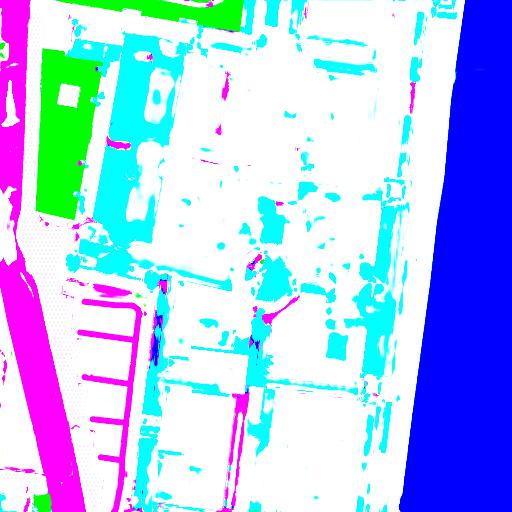}
    \end{subfigure}
    \begin{subfigure}[b]{0.23\textwidth}
        \centering
        \includegraphics[width=\textwidth]{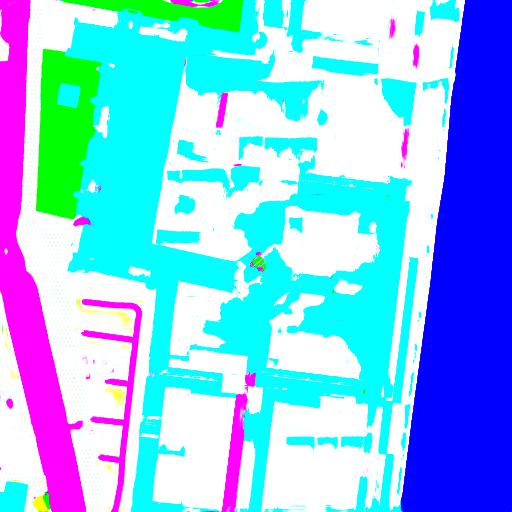}
    \end{subfigure}

    \vspace{4pt}

    \begin{subfigure}[b]{0.23\textwidth}
        \centering
        \includegraphics[width=\textwidth]{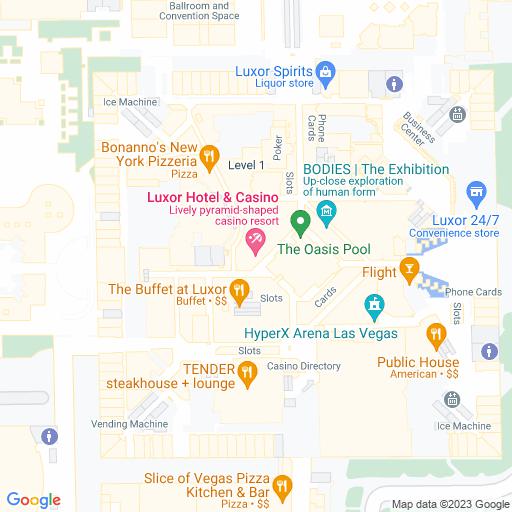}
    \end{subfigure}
    \begin{subfigure}[b]{0.23\textwidth}
        \centering
        \includegraphics[width=\textwidth]{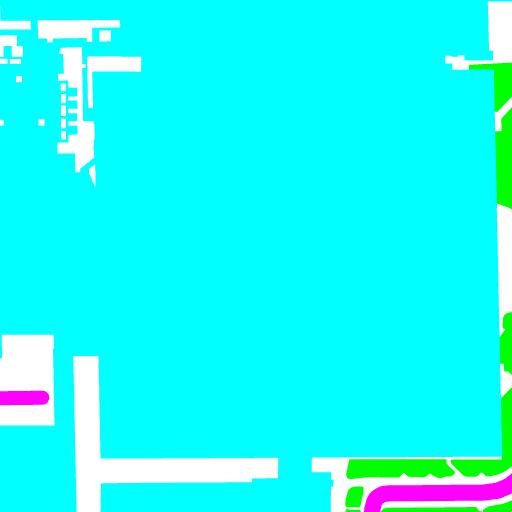}
    \end{subfigure}
    \begin{subfigure}[b]{0.23\textwidth}
        \centering
        \includegraphics[width=\textwidth]{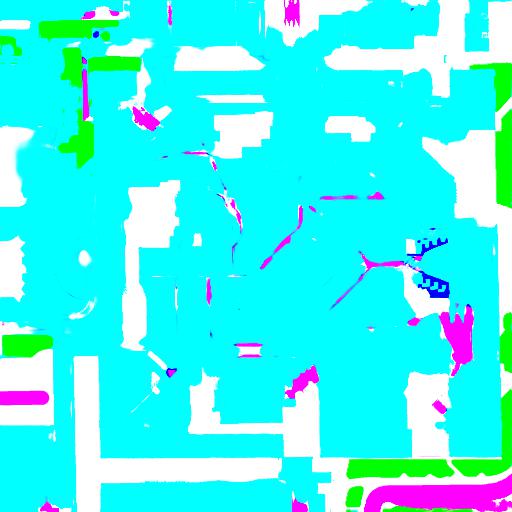}
    \end{subfigure}
    \begin{subfigure}[b]{0.23\textwidth}
        \centering
        \includegraphics[width=\textwidth]{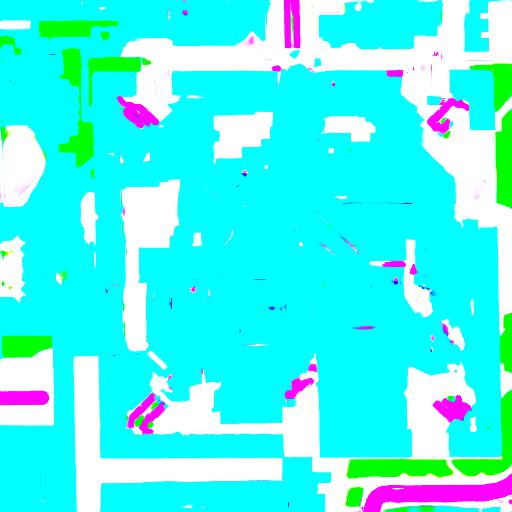}
    \end{subfigure}

    \begin{subfigure}[b]{0.23\textwidth}
        \centering
        \includegraphics[width=\textwidth]{Filler_horizontal.jpg}
        \caption{Source Map}
        \label{fig:results_src4}
    \end{subfigure}
    \begin{subfigure}[b]{0.23\textwidth}
        \centering
        \includegraphics[width=\textwidth]{Google_copyright_logo.jpg}
        \caption{Ground Truth}
        \label{fig:results_gt4}
    \end{subfigure}
    \begin{subfigure}[b]{0.23\textwidth}
        \centering
        \includegraphics[width=\textwidth]{Filler_horizontal.jpg}
        \caption{Double Zoom Model}
        \label{fig:results_double4}
    \end{subfigure}
    \begin{subfigure}[b]{0.23\textwidth}
        \centering
        \includegraphics[width=\textwidth]{Filler_horizontal.jpg}
        \caption{Single Zoom Model}
        \label{fig:results_single4}
    \end{subfigure}
    \caption{Additional model outputs illustrating limitations of the double and single zoom models.}
    \label{fig:additional_bad}
\end{figure}

Figure \ref{fig:additional_bad} gives additional outputs illustrating limitations of the double and single zoom models. Both models struggle with non-white backgrounds, 3D effects from buildings (seen at zooms of 18 and above), as well as mall interiors (also visible at zooms 18 and above). We hypothesize that these issues can be addressed with additional data augmentation approaches.

\begin{table}
\centering
\resizebox{\textwidth}{!}{
    \begin{tabular}{cc||ccc|ccc|ccc|ccc}
    \toprule
    & \textbf{Metric} & \multicolumn{3}{c|}{IoU} & \multicolumn{3}{c|}{F1} & \multicolumn{3}{c|}{Precision} & \multicolumn{3}{c}{Recall} \\
    & \textbf{Model} & Double & Single & \emph{Diff} & Double & Single & \emph{Diff} & Double & Single & \emph{Diff} & Double & Single & \emph{Diff} \\
    \hline
    \textbf{Class} & \textbf{Statistic} &  &  &  &  &  &  &  &  &  &  &  &  \\
    \midrule
    \multirow[c]{2}{*}{Streets} & median & 87.6 & 89.4 & -1.8 & 93.4 & 94.4 & -1.0 & 91.7 & 92.8 & -1.1 & 95.6 & 96.4 & -0.8 \\
     & mean & 86.8 & 88.9 & -2.1 & 92.8 & 94.1 & -1.3 & 91.5 & 92.5 & -1.0 & 94.4 & 95.8 & -1.4 \\
    \multirow[c]{2}{*}{Highways} & median & 87.4 & 88.5 & -1.1 & 93.3 & 93.9 & -0.6 & 90.2 & 89.9 & 0.3 & 97.4 & 98.9 & -1.5 \\
     & mean & 86.3 & 86.6 & -0.3 & 92.5 & 92.4 & 0.1 & 89.4 & 88.5 & 0.9 & 96.2 & 97.5 & -1.3 \\
    \multirow[c]{2}{*}{Parks} & median & 92.9 & 94.3 & -1.4 & 96.3 & 97.1 & -0.8 & 96.6 & 97.1 & -0.5 & 96.6 & 97.5 & -0.9 \\
     & mean & 89.2 & 91.0 & -1.8 & 93.3 & 94.4 & -1.1 & 94.0 & 94.9 & -0.9 & 94.0 & 95.2 & -1.2 \\
    \multirow[c]{2}{*}{Water} & median & 95.1 & 96.0 & -0.9 & 97.5 & 98.0 & -0.5 & 96.6 & 97.3 & -0.7 & 98.7 & 98.8 & -0.1 \\
     & mean & 91.7 & 93.4 & -1.7 & 95.4 & 96.4 & -1.0 & 93.8 & 95.7 & -1.9 & 97.3 & 97.3 & 0.0 \\
    \multirow[c]{2}{*}{Buildings} & median & N/A & N/A & N/A & N/A & N/A & N/A & N/A & N/A & N/A & N/A & N/A & N/A \\
     & mean & N/A & N/A & N/A & N/A & N/A & N/A & N/A & N/A & N/A & N/A & N/A & N/A \\
    \multirow[c]{2}{*}{Hospitals} & median & 79.8 & 79.9 & -0.1 & 88.7 & 88.8 & -0.1 & 89.2 & 89.8 & -0.6 & 88.3 & 87.4 & 0.9 \\
     & mean & 73.3 & 73.7 & -0.4 & 82.6 & 82.6 & 0.0 & 84.8 & 85.6 & -0.8 & 81.4 & 81.0 & 0.4 \\
     \bottomrule
    \end{tabular}
}
\caption{Class performances on the \textit{World} zoom 15 test set.}
\label{tab:zoom_15_ne_results}
\end{table}

\begin{table}
\centering
\resizebox{\textwidth}{!}{
    \begin{tabular}{cc||ccc|ccc|ccc|ccc}
    \toprule
    & \textbf{Metric} & \multicolumn{3}{c|}{IoU} & \multicolumn{3}{c|}{F1} & \multicolumn{3}{c|}{Precision} & \multicolumn{3}{c}{Recall} \\
    & \textbf{Model} & Double & Single & \emph{Diff} & Double & Single & \emph{Diff} & Double & Single & \emph{Diff} & Double & Single & \emph{Diff} \\
    \hline
    \textbf{Class} & \textbf{Statistic} &  &  &  &  &  &  &  &  &  &  &  &  \\
    \midrule
    \multirow[c]{2}{*}{Streets} & median & 94.8 & 96.3 & -1.5 & 97.3 & 98.1 & -0.8 & 96.5 & 97.6 & -1.1 & 98.3 & 98.7 & -0.4 \\
     & mean & 94.1 & 95.7 & -1.6 & 96.9 & 97.7 & -0.8 & 96.1 & 97.2 & -1.1 & 97.7 & 98.3 & -0.6 \\
    \multirow[c]{2}{*}{Highways} & median & 95.6 & 96.2 & -0.6 & 97.7 & 98.0 & -0.3 & 97.1 & 97.2 & -0.1 & 98.8 & 99.2 & -0.4 \\
     & mean & 94.6 & 94.8 & -0.2 & 97.2 & 97.2 & 0.0 & 96.6 & 96.4 & 0.2 & 97.9 & 98.3 & -0.4 \\
    \multirow[c]{2}{*}{Parks} & median & 96.8 & 97.6 & -0.8 & 98.4 & 98.8 & -0.4 & 98.7 & 99.0 & -0.3 & 98.2 & 98.8 & -0.6 \\
     & mean & 93.8 & 94.6 & -0.8 & 95.9 & 96.2 & -0.3 & 96.7 & 96.8 & -0.1 & 95.7 & 96.5 & -0.8 \\
    \multirow[c]{2}{*}{Water} & median & 97.3 & 98.2 & -0.9 & 98.6 & 99.1 & -0.5 & 98.7 & 98.8 & -0.1 & 99.1 & 99.3 & -0.2 \\
     & mean & 95.7 & 96.5 & -0.8 & 97.7 & 98.2 & -0.5 & 97.6 & 97.9 & -0.3 & 97.9 & 98.5 & -0.6 \\
    \multirow[c]{2}{*}{Buildings} & median & 0.1 & 0.0 & 0.1 & 0.2 & 0.0 & 0.2 & 2.3 & 0.0 & 2.3 & 0.1 & 0.0 & 0.1 \\
     & mean & 0.1 & 0.0 & 0.1 & 0.2 & 0.0 & 0.2 & 2.3 & 0.0 & 2.3 & 0.1 & 0.0 & 0.1 \\
    \multirow[c]{2}{*}{Hospitals} & median & 94.8 & 95.0 & -0.2 & 97.3 & 97.4 & -0.1 & 96.8 & 97.3 & -0.5 & 98.2 & 98.5 & -0.3 \\
     & mean & 91.7 & 92.4 & -0.7 & 94.8 & 95.2 & -0.4 & 95.6 & 96.0 & -0.4 & 95.2 & 95.7 & -0.5 \\
     \bottomrule
    \end{tabular}
}
\caption{Class performances on the \textit{World} zoom 16 test set.}
\label{tab:zoom_16_ne_results}
\end{table}

\begin{table}
\centering
\resizebox{\textwidth}{!}{
    \begin{tabular}{cc||ccc|ccc|ccc|ccc}
    \toprule
    & \textbf{Metric} & \multicolumn{3}{c|}{IoU} & \multicolumn{3}{c|}{F1} & \multicolumn{3}{c|}{Precision} & \multicolumn{3}{c}{Recall} \\
    & \textbf{Model} & Double & Single & \emph{Diff} & Double & Single & \emph{Diff} & Double & Single & \emph{Diff} & Double & Single & \emph{Diff} \\
    \hline
    \textbf{Class} & \textbf{Statistic} &  &  &  &  &  &  &  &  &  &  &  &  \\
    \midrule
    \multirow[c]{2}{*}{Streets} & median & 96.1 & 95.4 & 0.7 & 98.0 & 97.6 & 0.4 & 97.4 & 97.9 & -0.5 & 99.0 & 97.7 & 1.3 \\
     & mean & 94.6 & 93.3 & 1.3 & 97.1 & 96.4 & 0.7 & 96.3 & 97.3 & -1.0 & 98.1 & 95.8 & 2.3 \\
    \multirow[c]{2}{*}{Highways} & median & 97.8 & 97.0 & 0.8 & 98.9 & 98.5 & 0.4 & 98.6 & 98.5 & 0.1 & 99.4 & 98.7 & 0.7 \\
     & mean & 95.8 & 94.5 & 1.3 & 97.6 & 97.0 & 0.6 & 98.1 & 98.1 & 0.0 & 97.7 & 96.4 & 1.3 \\
    \multirow[c]{2}{*}{Parks} & median & 98.4 & 98.0 & 0.4 & 99.2 & 99.0 & 0.2 & 99.2 & 99.3 & -0.1 & 99.4 & 99.1 & 0.3 \\
     & mean & 94.9 & 93.5 & 1.4 & 96.3 & 95.4 & 0.9 & 97.0 & 96.8 & 0.2 & 96.6 & 95.3 & 1.3 \\
    \multirow[c]{2}{*}{Water} & median & 99.0 & 98.2 & 0.8 & 99.5 & 99.1 & 0.4 & 99.4 & 99.5 & -0.1 & 99.7 & 99.2 & 0.5 \\
     & mean & 98.3 & 97.4 & 0.9 & 99.2 & 98.7 & 0.5 & 99.0 & 99.1 & -0.1 & 99.3 & 98.3 & 1.0 \\
    \multirow[c]{2}{*}{Buildings} & median & 94.4 & 94.0 & 0.4 & 97.1 & 96.9 & 0.2 & 97.2 & 96.4 & 0.8 & 97.5 & 98.1 & -0.6 \\
     & mean & 92.0 & 91.9 & 0.1 & 95.4 & 95.4 & 0.0 & 95.9 & 94.1 & 1.8 & 95.4 & 97.2 & -1.8 \\
    \multirow[c]{2}{*}{Hospitals} & median & 96.7 & 95.6 & 1.1 & 98.3 & 97.7 & 0.6 & 98.8 & 99.2 & -0.4 & 97.1 & 96.3 & 0.8 \\
     & mean & 88.7 & 60.5 & 28.2 & 93.3 & 65.6 & 27.7 & 95.9 & 91.8 & 4.1 & 91.7 & 62.5 & 29.2 \\
     \bottomrule
    \end{tabular}
}
\caption{Class performances on the \textit{World} zoom 18 test set.}
\label{tab:zoom_18_ne_results}
\end{table}

\end{document}